\documentclass{article} 
\usepackage{iclr2027_conference,times}

\usepackage{amsmath,amsfonts,bm}

\def\eqref#1{equation~\ref{#1}}

\def\1{\bm{1}}

\DeclareMathAlphabet{\mathsfit}{\encodingdefault}{\sfdefault}{m}{sl}
\SetMathAlphabet{\mathsfit}{bold}{\encodingdefault}{\sfdefault}{bx}{n}

\usepackage{url}
\usepackage{booktabs}
\usepackage{graphicx}
\usepackage{xcolor}
\usepackage{pifont}
\usepackage{subcaption}
\usepackage{wrapfig}
\usepackage{xspace}
\usepackage[table]{xcolor} 
\definecolor{oursrow}{HTML}{D0F2C9}

\usepackage{colortbl}
\usepackage{array}
\usepackage{graphicx}
\usepackage{amsmath}  
\usepackage{bm}
\usepackage{multirow}
\usepackage[normalem]{ulem}
\usepackage[most]{tcolorbox}
\usepackage{wrapfig} 
\usepackage{bbding}
\usepackage{pifont}
\usepackage{enumitem}
\usepackage{adjustbox}
\usepackage{makecell}
\usepackage{tabularx}
\usepackage{array}
\newcolumntype{Y}{>{\centering\arraybackslash}X}
\usepackage{amssymb}
\usepackage[linesnumbered,ruled,vlined]{algorithm2e} 
\usepackage{algpseudocode} 
\usepackage{amsmath}
\usepackage{amsfonts}
\usepackage{bm}
\usepackage{setspace}
\usepackage{mathtools}
\usepackage{tablefootnote}
\usepackage[T1]{fontenc}
\usepackage{listings}
\usepackage{bbm}

\usepackage{minitoc}

\noptcrule

\definecolor{mycitecolor}{RGB}{1,103,204}  
\definecolor{tabletitle}{HTML}{DAD9E4}
\definecolor{table-blue}{RGB}{173, 216, 230}
\definecolor{row-highlight}{RGB}{220, 230, 242}
\definecolor{qwen-header}{RGB}{235, 240, 248}
\definecolor{llama-header}{RGB}{248, 244, 235}
\definecolor{section-gray}{RGB}{245, 245, 245}
\definecolor{group-header}{RGB}{230, 235, 245}  
\definecolor{groupcolor}{HTML}{F2F2F2}
\definecolor{ourscolor}{HTML}{FFF8E1}
\definecolor{avgcolor}{HTML}{F7FBFF}
\definecolor{githublink}{RGB}{251,49,153}
\definecolor{avgYellow}{RGB}{255,245,204}

\definecolor{linkgreen}{HTML}{3C7E7F}
\definecolor{rebuttal}{RGB}{255,0,0} 

\definecolor{zhz_gray}{rgb}{0.8,0.8,0.8}
\definecolor{darkgreen}{rgb}{0.0, 0.5, 0.0} 
\definecolor{darkred}{rgb}{0.5, 0.0, 0.0}   

\usepackage[colorlinks=true,
            citecolor=mycitecolor,   
            linkcolor=red,         
            urlcolor=githublink]          
           {hyperref}

\usepackage{tikz}
\usetikzlibrary{arrows.meta,positioning,fit,calc,shapes.misc,decorations.pathmorphing,backgrounds}

\title{\method{}: Adaptive Parametric Knowledge Participation for Faithful Retrieval-Augmented Generation}

\author{%
Ruochang Li$^{1,*}$,
Pengcheng Huang$^{1,*}$,
Zhenghao Liu$^{1,\dagger}$,
Yukun Yan$^{2}$ \\
\bfseries
Huiyuan Xie$^{2}$,
Yu Gu$^{1}$,
Ge Yu$^{1}$,
Maosong Sun$^{2}$ \\[4pt]
\mdseries
$^{1}$School of Computer Science and Engineering, Northeastern University, Shenyang, China \\
$^{2}$Department of Computer Science and Technology, Tsinghua University, Beijing, China \\[2pt]
}

\usepackage{xspace}
\def\method{{\fontfamily{lmtt}\selectfont \textbf{SPARK}}\xspace}

\iclrfinalcopy

\begin{document}

\maketitle
\lhead{Arxiv Preprint.}
{\renewcommand{\thefootnote}{\ensuremath{*}}\footnotetext{Equal contribution.\quad$^{\dagger}$Corresponding author.}}


\begin{abstract}

Retrieval-augmented generation (RAG) enhances the faithfulness of large language models (LLMs) by incorporating external knowledge into the generation process. However, conflicts between retrieved context and parametric knowledge can cause LLMs to rely on inconsistent internal knowledge, hindering their ability to faithfully leverage retrieved evidence. Existing approaches mitigate this issue by editing parametric knowledge to resolve such conflicts. However, locating and modifying knowledge encoded in model parameters is challenging and may inadvertently compromise the backbone LLM's capabilities.
In this paper, we introduce \method{}, a novel framework that formulates parametric knowledge control as a latent participation policy learning problem, enabling LLMs to adaptively regulate the contribution of parametric knowledge during generation.
Specifically, \method{} introduces a lightweight participation policy that dynamically modulates the contribution of FFN-mediated parametric knowledge while keeping the backbone LLM frozen. The participation policy is optimized with a faithfulness-oriented reward, enabling fine-grained control over parametric knowledge participation during generation.
Extensive experiments on six datasets validate the effectiveness of \method{}, which introduces and optimizes only 0.01\% additional trainable parameters while keeping the backbone LLM fully frozen, thereby preserving its general capabilities and achieving more faithful generation than competitive baselines.
All datasets and code are available at
\textcolor{githublink}{\url{https://github.com/OpenBMB/SPARK}}.

\end{abstract}

\section{Introduction}

Retrieval-augmented generation (RAG) has become a critical paradigm for improving the factual reliability of Large Language Models (LLMs) by grounding generation in external evidence~\citep{guu2020retrieval,lewis2020retrieval, izacard2021leveraging,li2025agenticragdeepreasoning,zhang2025websearchagenticdeep}.
By incorporating retrieved context into the generation process, RAG enables LLMs to access up-to-date and domain-specific information beyond the static parametric knowledge encoded during pretraining~\citep{izacard2023atlas, mallen-etal-2023-trust}.
A RAG system thus relies on two knowledge sources during inference: \textit{parametric knowledge}, which is encoded in model parameters, and \textit{contextual knowledge}, which is supplied by retrieved documents.
When both knowledge sources contradict each other, \textit{knowledge conflict} arises~\citep{xie2024adaptive}, under which LLMs may ignore retrieved evidence~\citep{zhang2025faithfulragfactlevelconflictmodeling}, overly rely on parametric knowledge~\citep{huang2025parammutesuppressingknowledgecriticalffns}, or inconsistently blend the two knowledge sources during generation~\citep{choi-etal-2025-conflict}.
These failure modes compromise the factual reliability of RAG systems~\citep{sun2025redeep}, making knowledge conflict mitigation a crucial problem for faithful retrieval-augmented generation.

A central challenge in mitigating knowledge conflict is how to regulate the influence of parametric knowledge when it contradicts retrieved evidence~\citep{longpre-etal-2021-entity,chen2022rich}.
Prior approaches typically follow a \emph{localize-then-intervene} paradigm: they first identify knowledge-related components and then apply predefined modifications, routing, or suppression strategies to regulate their contribution. 
However, knowledge localization does not necessarily identify the locations where intervention is most effective~\citep{hase2023doeslocalizationinformediting}, while predefined intervention strategies may fail to accommodate the evolving generation process~\citep{chen2025knowledgelocalizationmissionaccomplished}.
These limitations motivate us to shift our focus from identifying \emph{where} parametric knowledge should be intervened on toward understanding \emph{how} it should participate in generation under knowledge conflict.
%

In this paper, we propose \method{} (\textbf{S}elective \textbf{P}articipation of p\textbf{AR}ametric \textbf{K}nowledge), which recasts knowledge-conflict mitigation as learning how parametric knowledge should participate during generation.
Rather than relying on predefined intervention targets or fixed intervention rules, \method{} models the participation behavior of parametric knowledge during generation as a latent policy.
Since such participation behavior is inherently unobservable and cannot be directly annotated, the policy cannot be learned from explicit participation supervision.
We therefore keep the backbone LLM frozen and optimize only the participation policy with Group Relative Policy Optimization (GRPO)~\citep{guo2025deepseek}, using the faithfulness of generated responses as the learning signal.
In this way, \method{} learns latent knowledge participation directly from generation-level feedback, without requiring explicit labels for how parametric knowledge should participate.

Our experiments demonstrate the effectiveness of \method{} in improving the faithfulness of LLM generation, with \method{} outperforming competitive baselines by up to 3.49\% on standard RAG tasks. With fewer than 0.01\% additional trainable parameters, \method{} effectively modulates LLMs to appropriately leverage contextual knowledge while preserving their general capabilities.
Notably, \method{} delivers consistent improvements across diverse knowledge conflict scenarios, demonstrating its ability to learn tailored yet generalizable parametric knowledge participation policies for off-the-shelf LLMs. These policies enable LLMs to adaptively regulate the contribution of parametric knowledge across different scenarios, thereby improving generation faithfulness.
Moreover, we investigate the internal activation patterns of LLMs and find that the learned participation policies reshape neuron activations in ways that facilitate knowledge conflict resolution.


\section{Related Work}

Existing work on faithful retrieval-augmented generation has extensively investigated how to mitigate knowledge conflicts when contextual evidence contradicts a model's parametric memory
~\citep{longpre2022entitybasedknowledgeconflictsquestion,wang2025continuouslysteeringllmssensitivity}.
One line of research improves faithfulness by regulating how models incorporate contextual evidence in the presence of such conflicts.
For example, prompting-based methods encourage models to prioritize contextual evidence through carefully designed instructions~\citep{zhou2023context} or steer generation by contrasting predictions with and without contextual information~\citep{shi2024trusting}.
Another line optimizes context-following behavior through retrieval-augmented fine-tuning or preference optimization, encouraging models to develop a stronger reliance on external evidence~\citep{lin2024ra,bi2025context}.
More recently, adaptive decoding methods have shown that the relative contribution of contextual and parametric knowledge can vary across generation states.
For example, AdaCAD~\citep{wang2025adacadadaptivelydecodingbalance} dynamically adjusts the strength of contrastive decoding according to the degree of knowledge conflict, while CoCoA~\citep{khandelwal-etal-2025-cocoa} performs token-level adaptation based on contextual and parametric prediction signals. Despite these advances, existing approaches primarily regulate how external evidence is incorporated or how the output distribution is adjusted, while largely treating the model's internal parametric knowledge as a fixed source. How to selectively control the participation of internal knowledge during generation, particularly at the level of the model's underlying knowledge storage, remains largely unexplored.

A complementary line of work investigates the internal mechanisms through which parametric knowledge is stored and expressed.
To understand and enable targeted control over parametric knowledge, early studies attempted to localize factual knowledge at the neuron level~\citep{meng2023locatingeditingfactualassociations,dai2022knowledge}.
However, accurately localizing such knowledge remains challenging~\citep{geva-etal-2023-dissecting,chen2024journey}, as the resulting interventions are often computationally costly and brittle~\citep{hoelscher2023detecting,cohen2024evaluating,niu2024doesknowledgeneuronthesis}. 
Consequently, recent studies have shifted toward coarser-grained internal structures, such as attention heads and FFN modules, which provide more structured interfaces for knowledge control.
Building upon these mechanistic insights, several approaches directly intervene on internal components to mitigate knowledge conflicts.
For example, PH3~\citep{jin2024cutting}, RHIO~\citep{huang2025improving}, and JuICE~\citep{li2025taming} regulate generation through head-level pruning, contrastive decoding, or test-time intervention, while ROME~\citep{meng2023locatingeditingfactualassociations} directly edits FFN-based factual associations.
More recently, DoLa~\citep{chuang2024doladecodingcontrastinglayers} identifies intermediate layers and contrasts their logits with those from the final layer, without modifying model parameters. ParamMute~\citep{huang2025parammutesuppressingknowledgecriticalffns} identifies FFN modules associated with unfaithful generation and suppresses their contributions, while CoRect~\citep{ma2026corectcontextawarelogitcontrast} dynamically identifies suppressive FFN layers and performs hidden-state rectification
during inference.
Although these methods show that internal components can serve as effective control interfaces, they generally follow a \emph{localize-then-intervene} paradigm: they first identify components associated with undesirable parametric influence and then apply intervention strategies based on these identified components.
Furthermore, accurately identifying such components poses significant difficulties, as localization results can vary depending on the input and localization criteria~\citep{geva-etal-2023-dissecting,chen2024journey,chang-etal-2024-localization,chen2025knowledgelocalizationmissionaccomplished}.
More importantly, knowing \emph{where} knowledge is encoded does not by itself determine \emph{how much} the corresponding components should contribute during generation~\citep{hase2023doeslocalizationinformediting,chen2025knowledgelocalizationmissionaccomplished}. In contrast, \method{} reframes knowledge control as a \emph{participation} problem: rather than explicitly localizing and intervening on knowledge-bearing components, we learn a latent participation policy that dynamically determines how much each FFN contributes conditioned on the evolving generation context.

\begin{figure}[t]
    \centering
    \begin{subfigure}[t]{0.30\linewidth}
        \centering
        \includegraphics[width=\linewidth]{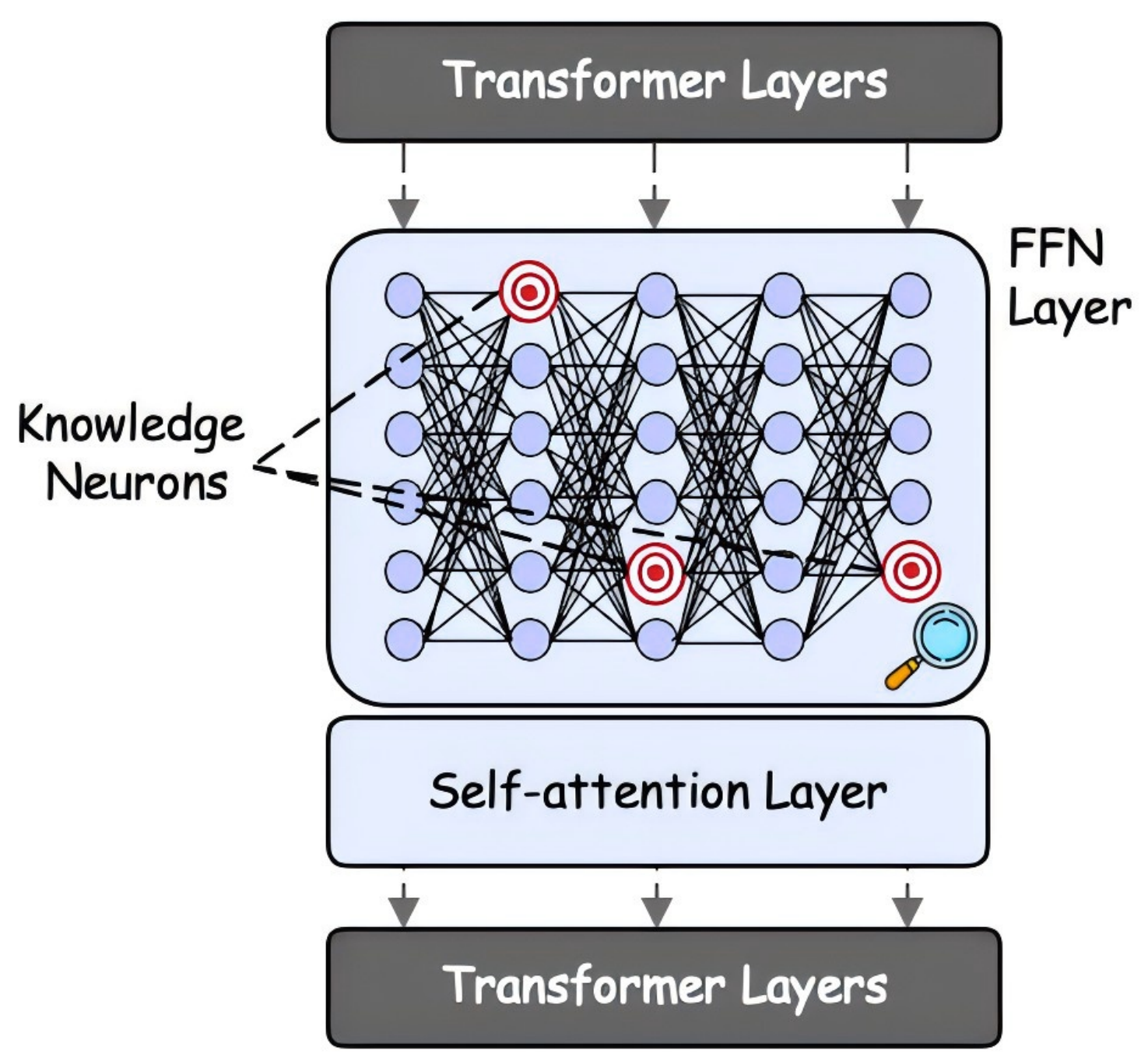}
        \caption{Neuron-level Intervention}
        \label{fig:method-a}
    \end{subfigure}
    \hfill
    \begin{subfigure}[t]{0.31\linewidth}
        \centering
        \includegraphics[width=\linewidth]{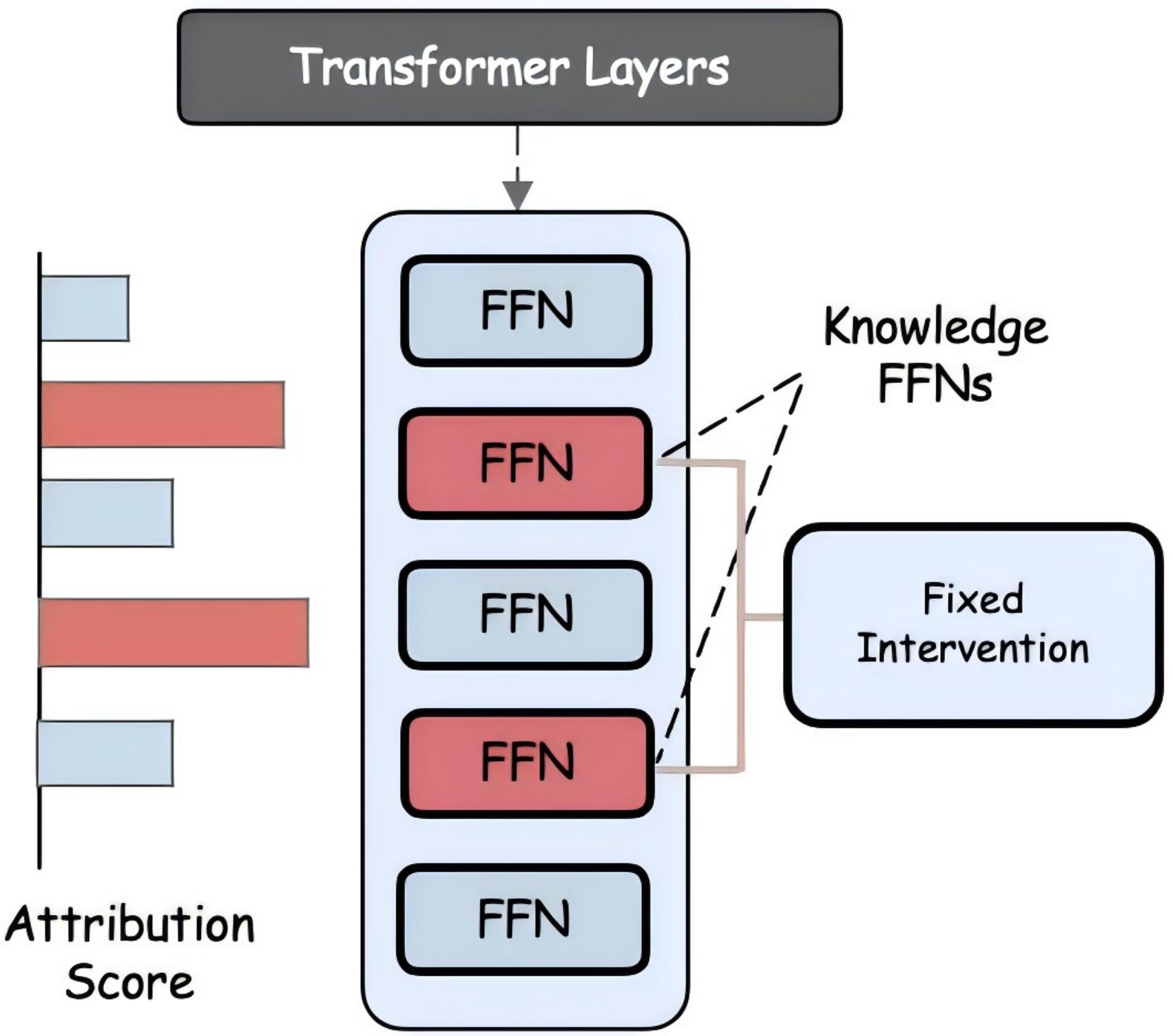}
        \caption{Layer-level Intervention}
        \label{fig:method-b}
    \end{subfigure}
    \hfill
    \begin{subfigure}[t]{0.37\linewidth}
        \centering
        \includegraphics[width=\linewidth]{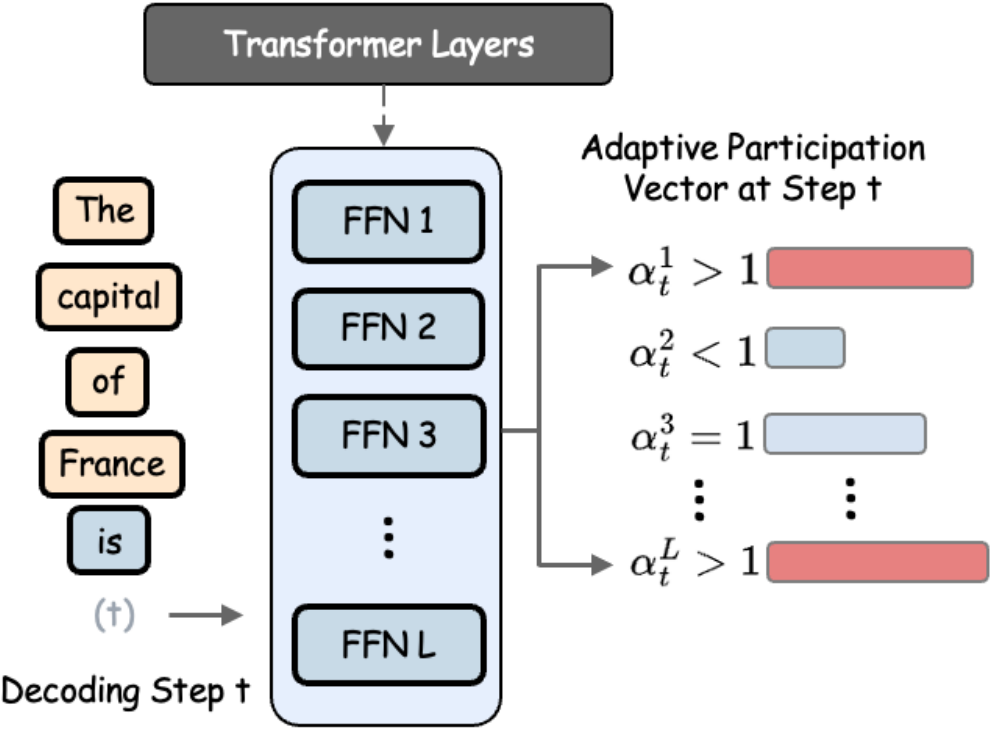}
        \caption{Our Method - \method{}}
        \label{fig:method-c}
    \end{subfigure}
    \vspace{-2mm}
    \caption{\textbf{Comparison of three paradigms for mitigating knowledge conflict.} (a) Neuron-level intervention requires fine-grained localization of knowledge-related neurons. (b) Layer-level intervention applies fixed rules to a manually selected subset of layers. (c) \method{} instead learns an input-dependent participation vector over FFN layers, thereby adapting how parametric knowledge contributes to generation.}
    \vspace{-1mm}
    \label{fig:method}
\end{figure}


\section{Preliminaries: From Knowledge Localization to Knowledge Participation in Faithful Generation}

In this work, we study how parametric knowledge should adaptively be expressed in generation to improve faithfulness under knowledge conflict.
Our key perspective is to move beyond \emph{knowledge localization} toward \emph{knowledge participation}: rather than merely identifying where knowledge is stored, we investigate whether and to what extent the corresponding components should participate in the generation process. In particular, we focus on feed-forward network (FFN) sublayers, which have been widely recognized as key components for storing and expressing parametric knowledge~\citep{geva2021transformerfeedforwardlayerskeyvalue,dai2022knowledge}. This makes FFNs a natural interface for controlling the contribution of parametric knowledge during generation.
We first motivate adaptive FFN participation by examining the limitations of predefined FFN-level interventions, which typically apply fixed modification strategies regardless of the evolving generation context. We then formulate FFN participation as a latent policy that dynamically determines whether and how FFN-mediated knowledge should contribute to generation, with the policy learned from response-level faithfulness feedback.


\subsection{Adaptive FFN Participation under Knowledge Conflict}
In Figure~\ref{fig:method}, we compare different levels of intervention for improving the faithfulness of LLM generation under knowledge conflict.

First, \emph{neuron-level intervention}, as illustrated in Figure~\ref{fig:method}(a), typically relies on attribution-based localization methods to identify knowledge neurons that are strongly associated with factual knowledge expression~\citep{dai-etal-2022-knowledge}. However, precisely localizing such neurons can be challenging and brittle, as factual knowledge is distributed across a large number of neurons~\citep{dai-etal-2022-knowledge}, while the identified neuron sets can vary substantially depending on the attribution method and the input query~\citep{chang-etal-2024-localization,chen2025knowledgelocalizationmissionaccomplished}, motivating operating at a coarser and more structured level of the model.
Recent studies therefore shift toward \emph{layer-level intervention} at the FFN-sublayer level, as shown in Figure~\ref{fig:method}(b). For example, existing approaches route between original and edited memories at selected FFNs or uniformly suppress a localized subset of knowledge-critical FFNs~\citep{wang2024wiserethinkingknowledgememory,huang2025parammutesuppressingknowledgecriticalffns}. These works demonstrate that FFN-level intervention provides a more robust interface for regulating the contribution of parametric knowledge. 
However, existing FFN-level approaches still follow a \emph{localize-then-intervene} paradigm: they first identify a predefined subset of knowledge-related FFNs and then apply fixed modification, routing, or suppression strategies to the selected components, regardless of the evolving generation context. 
Such predefined control not only restricts regulation to the localized subset, but may also impair the model's general capabilities by suppressing useful sublayers~\citep{hsueh-etal-2024-editing,fu-etal-2025-model}. 
These limitations motivate a shift from deciding \emph{where to intervene} toward determining \emph{when} and \emph{to what extent} each FFN should participate according to the current token representation.

As illustrated in Figure~\ref{fig:method}(c), our perspective reformulates parametric knowledge control as an \emph{adaptive participation problem}. Rather than committing to a predefined subset of localized FFNs, it considers all FFN sublayers as potential carriers of parametric knowledge within a functional subnetwork~\citep{zhang2022moeficationtransformerfeedforwardlayers}. This perspective models each FFN sublayer to exhibit an input-dependent participation pattern according to the evolving generation context.
Under this formulation, FFN participation can be represented as a latent policy over the degree to which each FFN contributes at each decoding step.


\subsection{Learning FFN Participation Policies via Reinforcement Learning}

The adaptive participation formulation introduces a latent policy learning problem because the optimal FFN participation coefficients are not directly observable, and reference responses do not provide explicit supervision for their values at individual FFN sublayers and token positions.
Therefore, the participation policy must be learned indirectly through its impact on the model's generation behavior.

After applying the FFN participation policy $\rho_{\psi}$ to the backbone LLM $\pi_{\theta}$, we obtain the $\rho_{\psi}$-augmented model $\pi_{\theta,\psi}$. During training, the backbone LLM $\pi_{\theta}$ remains frozen, while only the FFN participation policy $\rho_{\psi}$ is optimized. A straightforward way to learn this policy is through supervised fine-tuning (SFT):
\begin{equation}
\mathcal{L}_{\mathrm{SFT}}(\psi) =
- \sum_{t=1}^{T}
\log
\pi_{\theta,\psi}
\left(
y_t^{\star}
\mid
x,c,y_{<t}^{\star}
\right),
\end{equation}
where $y^{\star}$ is a reference response and parameters $\theta$ of the backbone LLM $\pi_\theta$ remain frozen. 
This objective indirectly optimizes the latent participation variables by maximizing the token-level likelihood of reference responses~\citep{ouyang2022traininglanguagemodelsfollow}.
However, it learns through next-token prediction under teacher-forced reference prefixes, which creates an objective mismatch with the response-level faithfulness desired for FFN participation.
Moreover, since every reference token serves as a supervised target, SFT is sensitive to the quality and diversity of the reference responses~\citep{li2025ragddroptimizingretrievalaugmentedgeneration,bukharin-etal-2024-data}.

To better align policy learning with response-level faithfulness, we instead optimize the FFN participation policy $\rho_{\psi}$ using response-level rewards.
Given the frozen backbone $\pi_{\theta}$, the participation policy $\rho_{\psi}$ induces a generation model $\pi_{\theta,\psi}$, whose objective is
\begin{equation}
\max_{\psi}\;
\mathbb{E}_{y\sim\pi_{\theta,\psi}(\cdot\mid x,c)}
\left[
R(y;x,c,y^*)
\right],
\end{equation}
where $R(y;x,c,y^*)$ provides response-level faithfulness feedback by
evaluating the generated response $y$ against the reference answer $y^*$.
Unlike SFT, this objective evaluates complete responses sampled from the model and directly uses response-level rewards to optimize the latent participation variables~\citep{shao2024deepseekmathpushinglimitsmathematical}. 
Reinforcement learning thus provides a natural way to learn latent FFN participation without requiring explicit supervision of participation labels~\citep{shao2024deepseekmathpushinglimitsmathematical}.



\section{Methodology}

\begin{figure}[t]
    \centering
    \includegraphics[width=\linewidth]{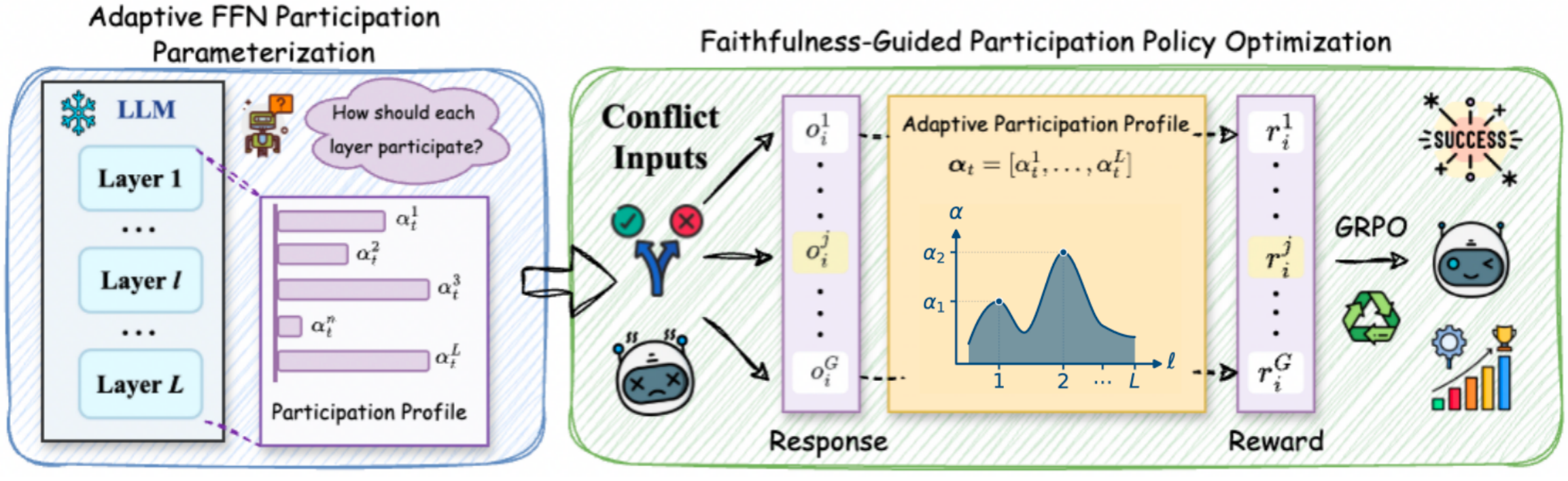}
    \vspace{-3mm}
    \caption{\textbf{Overview of the proposed framework \method{}.} The framework consists of two stages: (1) \emph{Adaptive FFN Participation Parameterization}, which introduces a layer-wise participation vector to dynamically regulate the contribution of parametric knowledge during generation; and (2) \emph{Faithfulness-Guided Participation Policy Optimization}, which learns the participation policy by optimizing response-level rewards that encourage faithful knowledge utilization.}
    \label{fig:framework}
\end{figure}

As illustrated in Figure~\ref{fig:framework}, this section presents \method{}, a framework that learns adaptive knowledge participation policies for faithful generation. 
The key idea is to formulate parametric knowledge participation as a latent policy learning problem during generation. 
We introduce a knowledge participation policy $\rho_\psi$, which observes the current generation state and produces layer-wise participation actions that determine the extent to which FFN-mediated parametric knowledge contributes to the generation process.
Specifically, \method{} first parameterizes the knowledge participation policy by introducing layer-wise participation modules into the LLM architecture (Section~\ref{sec:participation}). 
Then, the knowledge participation policy $\rho_{\psi}$ is optimized with faithfulness-oriented policy optimization, where generation-level rewards provide supervision for learning when and how parametric knowledge should participate during generation (Section~\ref{sec:grpo}).

\subsection{Parameterizing Adaptive FFN Participation in Off-the-Shelf LLMs}
\label{sec:participation}
Motivated by the observation that FFN sublayers serve as interfaces for expressing parametric knowledge, \method{} formulates knowledge participation as a learnable, input-dependent policy that determines how much FFN-mediated parametric knowledge should participate during generation.
Instead of modifying the parameters~\citep{huang2025parammutesuppressingknowledgecriticalffns}, we learn an adaptive participation policy that regulates the contribution of FFN-mediated parametric knowledge in off-the-shelf LLMs.

Specifically, the participation policy $\rho_{\psi}$ is parameterized by layer-wise participation modules. 
For each FFN layer $\ell$, where
$\ell\in\{1,\ldots, L\}$ and $L$ denotes the total number of FFN layers, the associated participation module has parameters $\psi_\ell=(\mathbf{w}_\ell,b_\ell)$. 
The complete policy parameters are denoted as
$\psi=\{\psi_\ell\}_{\ell=1}^{L}$. 
Given the hidden representation
$\mathbf{h}_{i,t}^{\ell}\in\mathbb{R}^{d}$ entering the FFN sublayer of layer $\ell$ at decoding step $t$ for the $i$-th input instance, we treat
$\mathbf{h}_{i,t}^{\ell}$ as the current generation state observed by the participation policy.
Formally, the participation policy produces a layer-wise participation action:
\begin{equation}
\alpha_{i,t}^{\ell}
=
\lambda_\alpha
\sigma
\left(
\mathbf{w}_{\ell}^{\top}
\mathbf{h}_{i,t}^{\ell}
+
b_{\ell}
\right),
\label{eq:factor}
\end{equation}
where $\sigma(\cdot)$ is the sigmoid function and $\alpha_{i,t}^{\ell}$ represents the participation action of the policy at layer $\ell$ and decoding step $t$.
A standard sigmoid output is bounded in $(0,1)$, allowing the participation policy only to attenuate FFN contributions. However, knowledge conflict does not necessarily indicate that parametric knowledge should be suppressed. When retrieved evidence is insufficient or unreliable, the model's parametric knowledge may instead provide useful complementary information, making stronger FFN contributions desirable. Thus, \method{} introduces a scaling factor $\lambda_\alpha$ that supports both attenuation and amplification of FFN contributions. This allows the participation policy to adaptively control the degree of parametric knowledge involvement during generation.

Accordingly, the participation action determines the extent to which
each FFN sublayer contributes to the residual stream:
\begin{equation}
    \mathbf{h}_{i,t}^{\ell+1}
    =
    \mathbf{h}_{i,t}^{\ell}
    +
    \alpha_{i,t}^{\ell}
    \operatorname{FFN}^{\ell}
    \left(
        \operatorname{Norm}
        (\mathbf{h}_{i,t}^{\ell})
    \right).
    \label{eq:participation_ffn}
\end{equation}

When $\alpha_{i,t}^{\ell}<1$, the corresponding FFN sublayer contributes less to the residual stream, reducing the participation of its associated
parametric knowledge.
Conversely, when $\alpha_{i,t}^{\ell}>1$, the FFN contribution is amplified, allowing internal knowledge to play a stronger role. 
Therefore, \method{} provides a continuous and input-dependent mechanism for adaptive FFN participation, rather than relying on predefined neuron selection or fixed layer-level intervention strategies.

To preserve the original generation behavior of the vanilla LLM before policy optimization, we initialize the parameters of the knowledge participation policy by setting $\mathbf{w}_{\ell}=0$ and initializing the bias as:
\begin{equation}
    b_{\ell}
    =
    \operatorname{logit}
    \left(
        \frac{1}{\lambda_\alpha}
    \right),
    \quad
    \forall \ell\in\{1,\ldots,L\}.
    \label{eq:participation_initialization}
\end{equation}
With this initialization, $\alpha_{i,t}^{\ell}=1$ for all hidden states, leaving the original FFN computation unchanged.
Therefore, the augmented model initially preserves the behavior of the frozen backbone LLM $\pi_\theta$, while subsequent optimization learns the adaptive participation policy $\rho_{\psi}$ through generation-level feedback (Section~\ref{sec:grpo}).

\subsection{Faithfulness-Guided Optimization of the Knowledge Participation Policy}
\label{sec:grpo}
Although existing training data $\mathcal{D}_{\mathrm{training}}$ provides factual correctness annotations for final responses, it lacks explicit supervision for the underlying knowledge participation decisions. 
In particular, the annotations do not specify when or to what extent parametric knowledge should participate during generation.
Since these fine-grained participation behaviors are latent, we formulate adaptive FFN participation as a policy optimization problem and optimize the participation policy $\rho_{\psi}$ using generation-level faithfulness feedback.

After incorporating the knowledge participation policy $\rho_{\psi}$ into the frozen backbone LLM $\pi_{\theta}$, we obtain the augmented generation model $\pi_{\theta,\psi}$. 
During optimization, we update only the parameters of the knowledge participation policy $\psi$, while keeping all backbone parameters $\theta$ of $\pi_{\theta}$ frozen.
The overall optimization objective is formulated as:
\begin{equation}
    \mathcal{L}
    =
    -\mathcal{J}_{\mathrm{GRPO}}
    +
    \beta_{\mathrm{ref}}\mathcal{L}_{\mathrm{ref}}
    + \lambda_{\mathrm{regular}}\mathcal{R}_{\mathrm{regular}},
    \label{eq:final_loss}
\end{equation}
where $\mathcal{J}_{\mathrm{GRPO}}$ is the GRPO objective~\citep{guo2025deepseek} that is used to optimize the knowledge participation policy according to faithfulness-oriented rewards, $\mathcal{L}_{\mathrm{ref}}$ constrains the optimized policy from deviating excessively from the original LLM behavior, and $\mathcal{R}_{\mathrm{regular}}$ regularizes the parameters of the learned participation policy $\rho_{\psi}$.
$\beta_{\mathrm{ref}}$ and $\lambda_{\mathrm{regular}}$ are hyperparameters.

Specifically, given an input instance
$(x_i,c_i,y_i^*)\sim\mathcal{D}_{\mathrm{training}}$,
where $x_i$ denotes the query, $c_i$ denotes the retrieved context, and
$y_i^*$ denotes the reference response, the augmented model
$\pi_{\theta,\psi}$ generates a group of $G$ candidate responses:
\begin{equation}
    \mathcal{O}_i
    =
    \{o_i^{1},o_i^{2},\ldots,o_i^{G}\},
    \qquad
    o_i^{j}
    \sim
    \pi_{\theta,\psi}
    (\cdot\mid x_i,c_i).
    \label{eq:group_generation}
\end{equation}
Each generated response $o_i^j$ is evaluated using a composite reward $R(o_i^j;x_i,c_i,y_i^*)$ comprising a format reward and a faithfulness reward, where the latter is computed by comparing the generated response with the reference answer $y_i^*$. More description is detailed in Appendix~\ref{appendix:method}.
GRPO then computes the relative advantages among sampled responses and updates the participation policy parameters $\psi$ to increase the probability of responses with higher faithfulness rewards.

To preserve the general generation capability of the original LLM, we adopt the frozen backbone model $\pi_{\theta}$ as the reference policy and introduce a KL-based regularization term:
\begin{equation}
    \mathcal{L}_{\mathrm{ref}}
    =
    \frac{1}{T}
    \sum_{t=1}^{T}
    D_{\mathrm{KL}}
    \left(
        \pi_{\theta,\psi}
        (\cdot|x_i,c_i,o_{i,<t})
        \|
        \pi_{\theta}
        (\cdot|x_i,c_i,o_{i,<t})
    \right),
    \label{eq:kl_ref}
\end{equation}
where the KL divergence is computed at each decoding step over the token distribution, and $T$ denotes the response length of the LLM response. 
This constraint encourages the learned knowledge participation policy $\rho_{\psi}$ to improve faithfulness while preserving the behavior of the original LLM.

Furthermore, since the participation policy is parameterized by layer-wise
FFN participation parameters, we impose an $L_2$ regularization term:
\begin{equation}
    \mathcal{R}_{\mathrm{regular}}
    =
    \frac{1}{L}
    \sum_{\ell=1}^{L}
    \left(
        \|\mathbf{w}_{\ell}\|^2
        +
        b_{\ell}^{2}
    \right).
    \label{eq:participation_reg}
\end{equation}
This regularization prevents overly aggressive participation adjustments and encourages the learned policy to achieve faithful generation through adaptive FFN participation.

\section{Experimental Methodology}
This section first describes datasets, evaluation metrics, and baselines. Then, we introduce the implementation details of our experiments. More experimental details are shown in Appendix~\ref{appendix:hyperparameters}.

\label{sec:setup}

\begin{table*}[t!]
\centering
\vspace{-0.1em}
\caption{\textbf{Performance comparison on the MRQA benchmark.}
The best and second-best results are marked in \textbf{bold} and \underline{underlined}, respectively. $^{\dag}$ and $^{\ddag}$ indicate statistically significant improvements over SFT$^{\dag}$ and Knowledgeable-R1$^{\ddag}$ in F1 score.}
\scriptsize
\setlength{\tabcolsep}{4.5pt}
\renewcommand{\arraystretch}{1.15}
\resizebox{\textwidth}{!}{
\begin{tabular}{l|ccccccc}
\Xhline{1.2pt}
\rowcolor{tabletitle}
\textbf{Methods}
& \textbf{HotpotQA}
& \textbf{SearchQA}
& \textbf{NewsQA}
& \textbf{NQ}
& \textbf{TriviaQA}
& \textbf{SQuAD}
& \textbf{Avg.} \\
\Xhline{1.2pt}

\multicolumn{8}{c}{\textit{Qwen-3-1.7B}} \\
\hline

\rowcolor{gray!10}
No-RAG~\citep{roberts2020much}
& 8.82
& 14.77
& 3.09
& 9.22
& 18.33
& 10.40
& 10.77 \\

Vanilla-RAG~\citep{lewis2020retrieval}
& 63.41
& 53.89
& 33.08
& 55.07
& 62.53
& 68.15
& 56.02 \\

\rowcolor{gray!10}
CtrlA~\citep{liu2024ctrlaadaptiveretrievalaugmentedgeneration}
& 24.39
& 30.12
& 12.92
& 22.49
& 34.54
& 23.78
& 24.71 \\

CK-PLUG~\citep{bi2025parametersvscontextfinegrained}
& 57.14
& 58.67
& \underline{51.31}
& 58.44
& 76.71
& 76.42
& 63.12 \\

\rowcolor{gray!10}
IRCAN~\citep{shi2024ircanmitigatingknowledgeconflicts}
& 52.62
& 49.76
& 23.19
& 40.42
& 50.91
& 53.64
& 45.09 \\

SFT~\citep{wei2021finetuned}
& \underline{65.81}
& 67.62
& 47.68
& \underline{63.83}
& 71.75
& \underline{76.52}
& \underline{65.53} \\

\rowcolor{gray!10}
Knowledgeable-r1~\citep{lin2026resistingcontextualinterferencerag}
& 59.75
& 61.86
& 41.90
& 59.11
& 73.05
& 77.08
& 62.13 \\

CoRect~\citep{ma2026corectcontextawarelogitcontrast}
& 65.01
& 54.54
& 45.93
& 59.45
& 67.14
& 72.96
& 60.84 \\

\textbf{\method{} (ours)}
& \textbf{69.30}$^{\dagger\ddagger}$
& \underline{63.86}$^{\ddagger}$
& \textbf{52.91}$^{\dagger\ddagger}$
& \textbf{65.55}$^{\dagger\ddagger}$
& \underline{74.17}$^{\dagger\ddagger}$
& 75.99
& \textbf{66.96}$^{\dagger\ddagger}$ \\

\hline
\multicolumn{8}{c}{\textit{Qwen-3-8B}} \\
\hline

\rowcolor{gray!10}
No-RAG~\citep{roberts2020much}
& 21.53
& 49.76
& 6.22
& 28.07
& 56.52
& 23.70
& 30.97 \\

Vanilla-RAG~\citep{lewis2020retrieval}
& 78.81
& 75.05
& 60.22
& \underline{70.99}
& 81.07
& 84.32
& 75.08 \\

\rowcolor{gray!10}
CtrlA~\citep{liu2024ctrlaadaptiveretrievalaugmentedgeneration}
& 47.28
& 61.01
& 43.52
& 52.55
& 66.41
& 59.05
& 54.97 \\

CK-PLUG~\citep{bi2025parametersvscontextfinegrained}
& 67.90
& 69.26
& 55.59
& 65.56
& 78.09
& 80.39
& 69.47 \\

\rowcolor{gray!10}
IRCAN~\citep{shi2024ircanmitigatingknowledgeconflicts}
& 78.71
& 74.35
& 58.34
& 68.73
& 81.19
& 83.52
& 74.14 \\

SFT~\citep{wei2021finetuned}
& 73.10
& 76.39
& 59.10
& 69.43
& 78.19
& 85.93
& 75.23 \\

\rowcolor{gray!10}
Knowledgeable-r1~\citep{lin2026resistingcontextualinterferencerag}
& \underline{78.84}
& 75.71
& \underline{60.46}
& 70.66
& \underline{81.78}
& 84.77
& \underline{75.37} \\

CoRect~\citep{ma2026corectcontextawarelogitcontrast}
& 57.19
& 49.15
& 56.25
& 58.52
& 68.45
& 76.49
& 61.01 \\

\textbf{\method{} (ours)}
& \textbf{79.08}$^{\dagger\ddagger}$
& \underline{76.27}$^{\ddagger}$
& \textbf{60.93}$^{\dagger\ddagger}$
& \textbf{71.57}$^{\dagger\ddagger}$
& \textbf{82.28}$^{\dagger\ddagger}$
& \underline{85.50}$^{\ddagger}$
& \textbf{75.94}$^{\dagger\ddagger}$ \\

\Xhline{1.2pt}
\end{tabular}
}
\label{tab:main_res}
\end{table*}

\begin{wraptable}{r}{0.53\textwidth}
\vspace{-0.8em}
\centering
\caption{\textbf{Performance comparison on the ConfiQA benchmark.}
The best and second-best results are marked in \textbf{bold} and
\underline{underlined}, respectively.}
\label{tab:confiqa_res}

\scriptsize
\setlength{\tabcolsep}{8pt}
\renewcommand{\arraystretch}{1.15}

\begin{tabular}{l|ccc}
\Xhline{1.2pt}
\rowcolor{tabletitle}
\textbf{Methods}
& \textbf{MC}
& \textbf{MR}
& \textbf{QA} \\
\Xhline{1.2pt}

\multicolumn{4}{c}{\textit{Qwen-3-1.7B}} \\
\hline

\rowcolor{gray!10}
Vanilla-RAG~\citep{lewis2020retrieval}
& 7.82
& 11.10
& 5.70 \\

SFT~\citep{wei2021finetuned}
& \underline{7.37}
& \underline{8.85}
& \underline{9.92} \\

\rowcolor{gray!10}
Knowledgeable-r1~\citep{lin2026resistingcontextualinterferencerag}
& 8.17
& 11.72
& 6.70 \\

\textbf{\method{} (ours)}
& \textbf{10.00}
& \textbf{14.20}
& \textbf{10.80} \\

\hline
\multicolumn{4}{c}{\textit{Qwen-3-8B}} \\
\hline

\rowcolor{gray!10}
Vanilla-RAG~\citep{lewis2020retrieval}
& \underline{18.45}
& \underline{21.34}
& \underline{27.17} \\

SFT~\citep{wei2021finetuned}
& 7.80
& 11.70
& 12.60 \\

\rowcolor{gray!10}
Knowledgeable-r1~\citep{lin2026resistingcontextualinterferencerag}
& 14.72
& 20.83
& 19.42 \\

\textbf{\method{} (ours)}
& \textbf{21.10}
& \textbf{30.30}
& \textbf{28.60} \\

\Xhline{1.2pt}
\end{tabular}

\vspace{-0.8em}
\end{wraptable}

\noindent\textbf{Datasets.}
To provide a comprehensive evaluation, we conduct extensive experiments on three categories of publicly available datasets: \textit{\ding{182} In-domain Evaluation:} We evaluate our method on the standard open-domain MRQA benchmark~\citep{fisch-etal-2019-mrqa}, including HotpotQA~\citep{yang-etal-2018-hotpotqa}, SearchQA~\citep{dunn2017searchqa}, NewsQA~\citep{trischler-etal-2017-newsqa}, Natural Questions (NQ)~\citep{kwiatkowski-etal-2019-natural}, TriviaQA~\citep{joshi-etal-2017-triviaqa}, and SQuAD~\citep{rajpurkar2016squad}.
\textit{\ding{183} Out-of-domain Evaluation:} We further evaluate our model on ConfiQA~\citep{bi2024contextdpoaligninglanguagemodels} to examine its generalization capabilities, including Question Answering (QA), Multi-hop Reasoning (MR), and Multi-Conflicts (MC). 
\textit{ \ding{184} General Capability Preservation:} We report performance on the MMLU benchmark~\citep{hendrycks2021measuringmassivemultitasklanguage} to verify the general capability preservation of our \method{}. 
Details are provided in Appendix~\ref{appendix:datasets}.

\noindent\textbf{Baselines.}
We evaluate our proposed \method{} against a range of competitive baselines, which can be categorized into three groups: \ding{182} \textit{Prompting-based methods}, including Vanilla-RAG~\citep{ram2023incontextretrievalaugmentedlanguagemodels}, CtrlA~\citep{liu2024ctrlaadaptiveretrievalaugmentedgeneration}, which performs adaptive retrieval by leveraging representation-level control signals.
\ding{183} \textit{Decoding-based methods}, including 
IRCAN~\citep{shi2024ircanmitigatingknowledgeconflicts}, which identifies and reweights context-aware neurons to strengthen contextual cues, 
CoRect~\citep{ma2026corectcontextawarelogitcontrast}, which performs training-free hidden-state rectification via context-aware logit contrast,
and CK-PLUG~\citep{bi2025parametersvscontextfinegrained}, which provides plug-and-play control over the model's reliance by modifying token probability distributions.\ding{184} \textit{Fine-tuning methods}, including Supervised Fine-Tuning (SFT)~\citep{wei2021finetuned} and Knowledgeable-R1~\citep{lin2026resistingcontextualinterferencerag}, which train the model to improve its behavior under contextual knowledge conflicts.
Detailed descriptions are provided in Appendix~\ref{appendix:baselines}.

\noindent\textbf{Evaluation Metrics.} 
Following~\citet{longpre-etal-2021-entity}, we evaluate the MRQA benchmark with token-level F1 score. 
To further evaluate the generalization of \method{}, we also adopt the ConfiQA benchmark with Exact Match (EM), following~\citet {bi2025context}.
In addition, we report 5-shot accuracy on MMLU to assess whether \method{} preserves general capabilities.

\noindent\textbf{Implementation Details.}
To ensure a fair comparison, we use Qwen-3-1.7B and Qwen-3-8B in non-thinking mode as the backbone models throughout our experiments.
For \method{}, we set the hyperparameters $\lambda_g$ to 2, which controls the modulation range in Eq.~\ref{eq:factor}.
We set both $\mathbf{w}_l$ and $b_l$ to 0 in Eq.~\ref{eq:participation_reg} for all layers.
Additional details, including training data construction, hyperparameter settings, and training prompt templates, are provided in Appendix~\ref{appendix:data_construction}, Appendix~\ref{appendix:hyperparameters}, and Appendix~\ref{appendix:prompt}.

\section{Experimental Analysis}
\label{sec:exp_analysis}

In this section, we first show the overall performance of \method{} and conduct ablation studies to show the contribution of different components. 
We then analyze the behavior of \method{} to explore how it adaptively regulates the participation of parametric knowledge under different knowledge conflict scenarios and investigate the resulting neuron activation patterns within the LLMs.
Finally, we present several case studies in Appendix~\ref{appendix:case}.

\begin{wraptable}{r}{0.55\textwidth}
\vspace{-0.8em}
\centering
\caption{
\textbf{Ablation study on the MRQA benchmark.}
The best and second-best results are marked in \textbf{bold} and
\underline{underlined}, respectively.
}
\label{tab:ablation_study}

\scriptsize
\setlength{\tabcolsep}{4.5pt}
\renewcommand{\arraystretch}{1.15}

\begin{tabular}{l|cccc}
\Xhline{1.2pt}
\rowcolor{tabletitle}
\textbf{Methods}
& \textbf{HotpotQA}
& \textbf{SearchQA}
& \textbf{TriviaQA}
& \textbf{Avg.} \\
\Xhline{1.2pt}

\multicolumn{5}{c}{\textit{Qwen-3-1.7B}} \\
\hline

\rowcolor{gray!10}
Gate-SFT
& 64.76
& \underline{61.80}
& 68.93
& 65.16 \\

Static-Gate GRPO
& 63.68
& 54.05
& 62.74
& 60.16 \\

\rowcolor{gray!10}
w/o $L_2$ Regulation
& 69.85
& 59.03
& \underline{73.98}
& \underline{67.62} \\

w/o Faithful Reward
& \underline{69.52}
& 54.22
& 71.85
& 65.20 \\

\rowcolor{gray!10}
w/o Format Reward
& 68.27
& 60.02
& 72.59
& 66.96 \\

\textbf{\method{}}
& 69.30
& \textbf{63.86}
& \textbf{74.17}
& \textbf{69.11} \\

\hline
\multicolumn{5}{c}{\textit{Qwen-3-8B}} \\
\hline

\rowcolor{gray!10}
Gate-SFT
& 76.95
& 78.29
& \underline{81.61}
& \underline{78.95} \\

Static-Gate GRPO
& 78.80
& 75.05
& 81.16
& 78.34 \\

\rowcolor{gray!10}
w/o $L_2$ Regulation
& \underline{79.08}
& 75.06
& 80.90
& 78.35 \\

w/o Faithful Reward
& 79.01
& 75.00
& 81.20
& 78.40 \\

\rowcolor{gray!10}
w/o Format Reward
& 78.92
& \underline{75.15}
& 81.05
& 78.37 \\

\textbf{\method{}}
& \textbf{79.08}
& 76.27
& \textbf{82.28}
& \textbf{79.21} \\

\Xhline{1.2pt}
\end{tabular}

\vspace{-0.8em}
\end{wraptable}

\subsection{Overall Performance}
\label{sec:main_res}

Table~\ref{tab:main_res} shows the performance of \method{} against the compared methods for the MRQA benchmark on Qwen3-1.7B and Qwen3-8B.
Overall, \method{} consistently achieves state-of-the-art performance, outperforming the strongest baseline by up to 3.49\%.
Compared with No-RAG, Vanilla RAG significantly enhances LLM performance across all datasets, indicating that external knowledge effectively improves the accuracy of generated responses.
The intervention methods such as CtrlA, IRCAN, CoRect, and CK-PLUG yield unstable gains, suggesting that predefined decoding interventions struggle to regulate parametric knowledge.
Specifically, IRCAN and CoRect underperform Vanilla RAG on Qwen3-1.7B, and CoRect remains substantially weaker on Qwen3-8B, indicating that training-free neuron or logit interventions do not reliably scale.
While fine-tuning methods are more competitive on the MRQA benchmark, SFT and Knowledgeable-R1 attain the strongest baseline on Qwen3-1.7B and Qwen3-8B, respectively.
Although the improvement over the strongest baseline becomes smaller on the larger backbone, \method{} remains consistently best on average, and its advantage is not confined to a single dataset.
Moreover, Table~\ref{tab:confiqa_res} presents the performance of \method{} on the ConfiQA benchmark to evaluate out-of-domain generalization.
\method{} consistently outperforms all compared methods across the MC, MR, and QA subsets on both backbones.
Overall, the consistent gains on both in-domain MRQA and out-of-domain ConfiQA indicate that \method{} learns an adaptive participation policy that regulates parametric knowledge according to the evolving generation context, rather than simply relying on either knowledge source.

\subsection{Ablation Studies}
\label{sec:ablation}
Table~\ref{tab:ablation_study} presents ablation studies on Qwen3-1.7B and Qwen3-8B to examine how different components contribute to learning an effective knowledge participation policy.
Specifically, we compare the full model with the following variants:
\ding{182} Gate-SFT, which keeps the same trainable gates but replaces GRPO with supervised fine-tuning.
\ding{183} Static-Gate GRPO, which retains GRPO but replaces the input-dependent participation vector with an input-independent bias.
\ding{184} w/o $L_2$ Regulation, which removes the regularization imposed on the knowledge participation policy.
\ding{185} w/o Faithful Reward, which removes the answer-faithfulness reward and retains only the format reward.
\ding{186} w/o Format Reward, which removes the format reward and retains only the faithfulness reward.
Overall, the results suggest that these components play complementary roles rather than merely providing additive gains.
Compared with Gate-SFT, \method{} improves average F1 by up to 3.95\% on Qwen3-1.7B, showing that reinforcement learning is necessary beyond training the gates with supervised objectives alone.
Replacing input-dependent gates with Static-Gate GRPO further leads to clear drops, especially on Qwen3-1.7B, indicating that adaptive, instance-conditioned participation is critical.
The faithfulness reward is particularly important for Qwen3-1.7B, suggesting that smaller models rely more on explicit response-level supervision to learn effective knowledge participation.
In particular, retaining only the format reward already yields competitive scores, yet reintroducing the faithfulness reward further improves average F1 by 3.91\% and 0.81\% on Qwen3-1.7B and Qwen3-8B, respectively, and this gap indicates that formatting supervision mainly stabilizes answer extraction, while the additional gains of \method{} come from faithful generation.
In contrast, $L_2$ regularization improves robustness across datasets by preventing overly aggressive participation adjustments, with a more pronounced effect on Qwen3-8B.
The consistent degradation after removing the format reward further indicates that structural supervision complements faithfulness-oriented optimization.
Overall, format reward is necessary, as it provides a reliable response structure, whereas faithfulness-oriented optimization is what drives the knowledge-participation improvements.

\begin{wrapfigure}{r}{0.5\textwidth}
    \centering
    \begin{subfigure}[t]{0.50\linewidth}
        \centering
        \includegraphics[width=\linewidth]{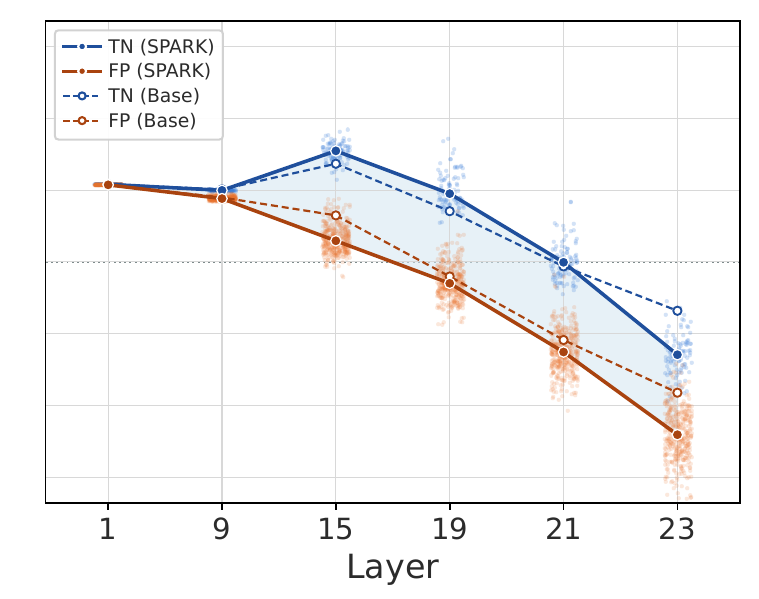}
        \caption{TN/FP scenario separation across layers of Qwen3-1.7B on NewsQA.}
        \label{fig:method-a}
    \end{subfigure}
    \hfill
    \begin{subfigure}[t]{0.46\linewidth}
        \centering
        \includegraphics[width=\linewidth]{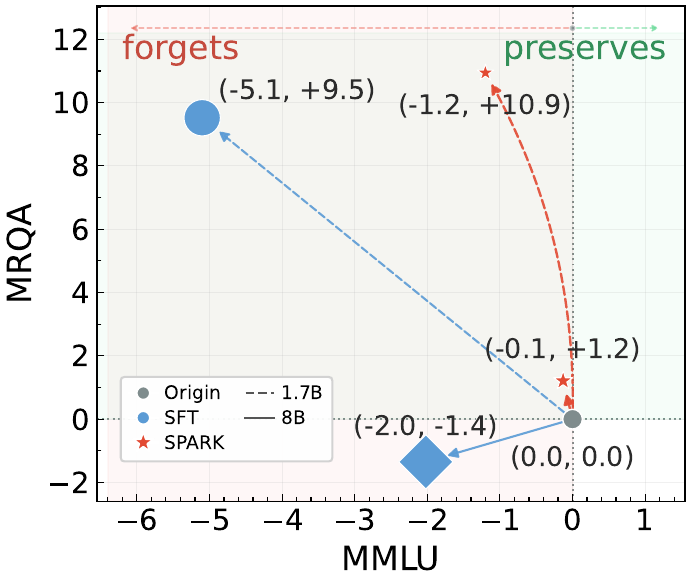}
        \caption{Performance on the MMLU and MRQA benchmarks.}
        \label{fig:method-b}
    \end{subfigure}
    \vspace{-2mm}
    \caption{\textbf{Characteristics of adaptive knowledge participation.} Marker size in Figure(a) indicates the number of trainable parameters.}
\label{fig:adaptive_char}
    \vspace{-1mm}
    \label{fig:last_fig}
\end{wrapfigure}

\subsection{Characteristics of Adaptive Knowledge Participation in \method{}}
\label{sec:characteristics}
In this experiment, we explore whether \method{} could yield consistent improvements across
diverse knowledge-conflict scenarios.
To quantify this process, we first define two typical knowledge conflict scenarios: \ding{182}~TN represents that the parametric answer is correct, but the context is wrong. \ding{183}~FP represents that the parametric answer is wrong and the context is correct.
We then evaluate our \method{} under each scenario and find that on both Qwen3-1.7B and Qwen3-8B, \method{} outperforms or matches the most competitive baseline across both scenarios on average, detailed in Appendix~\ref{appendix:scenarios}.
To further investigate why \method{} benefits all scenarios, following the linear representation hypothesis~\citep{park2024linearrepresentationhypothesisgeometry,belrose2025elicitinglatentpredictionstransformers}, we project per-sample residual-stream activations onto the Fisher discriminant direction at each layer. As shown in Figure~\ref{fig:last_fig}(a), 
compared with the base model, \method{} produces wider TN/FP separation from layer~15 onward, indicating that \method{}'s participation policy strengthens the model's internal differentiation between scenarios where its parametric knowledge is reliable versus unreliable, providing a clearer representational basis for faithful generation.
To complement this layer-wise analysis, we further provide a t-SNE visualization of per-sample participation vectors in Appendix~\ref{appendix:visiualization}, which shows that \method{} induces more clearly separated TN/FP clusters than the base model.
In addition to the scenario-specific improvements above, \method{}
also preserves general capabilities. To verify this, we further
evaluate on MMLU, and Figure~\ref{fig:last_fig}(b) shows that \method{} remains competitive with the
original model while training less than 0.004\% of the backbone parameters, whereas full parameter SFT drops up to 5.1.
More specifically, on Qwen3-1.7B, \method{} improves MRQA by 10.9 while sacrificing only 1.2 on MMLU, whereas full-parameter SFT gains 9.5 on MRQA at the cost of a 5.1 MMLU drop.
These results suggest that freezing the backbone and updating only a lightweight participation policy is sufficient to improve conflict handling, while largely avoiding the catastrophic forgetting typically induced by full-parameter fine-tuning.

\begin{wrapfigure}{r}{0.50\textwidth}
    \centering
    \begin{subfigure}[t]{0.47\linewidth}
        \centering
        \includegraphics[width=\linewidth]{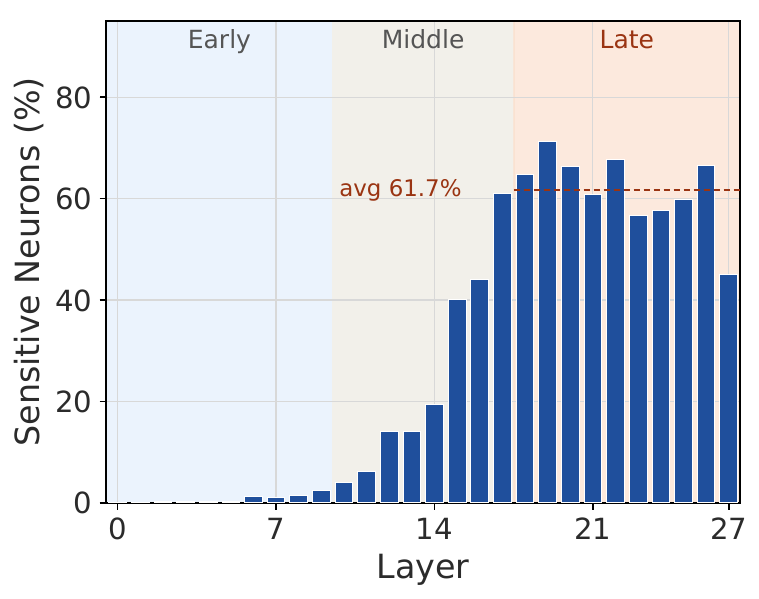}
        \caption{Layer-wise concentration of scenario-sensitive neurons on Qwen3-1.7B.}
        \label{fig:neuron-a}
    \end{subfigure}
    \hfill
    \begin{subfigure}[t]{0.46\linewidth}
        \centering
        \includegraphics[width=\linewidth]{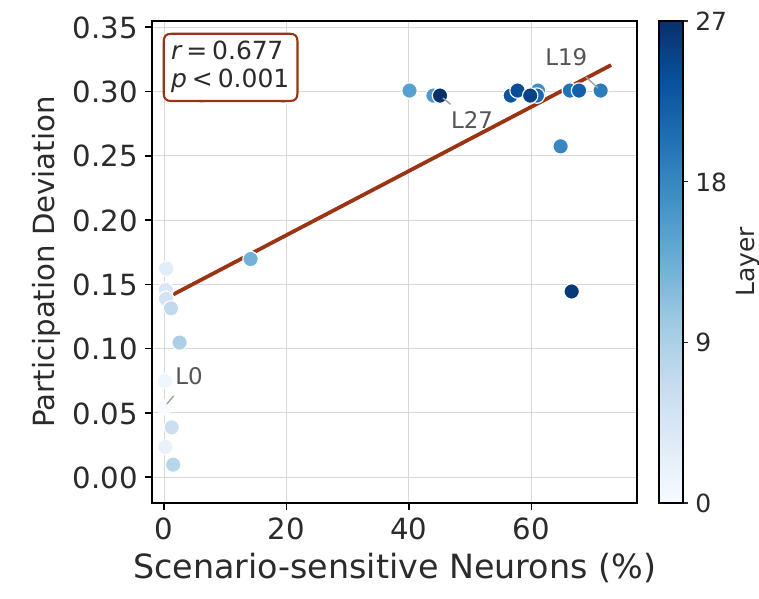}
        \caption{Participation-profile deviation vs.\ sensitive-neuron density.}
        \label{fig:neuron-b}
    \end{subfigure}
    \vspace{-2mm}
    \caption{\textbf{Neuron-level analysis under knowledge conflicts.}
    (a)~Sensitive neurons concentrate in late layers.
    (b)~Larger participation-profile deviation coincides with higher sensitive-neuron density.
    Point color in (b) encodes layer index.}
    \label{fig:neuron_sensitivity}
    \vspace{-1mm}
\end{wrapfigure}

\subsection{Analysis of Neuron Activation Patterns in \method{} under Knowledge Conflicts}
\label{sec:neuron_analysis}
In Section~\ref{sec:characteristics}, we have revealed that \method{}'s participation policy amplifies scenario separation at the residual-stream level. Inspired by MOEfication~\citep{zhang2022moeficationtransformerfeedforwardlayers}, which demonstrates that FFN layers can be decomposed into functionally specialized neuron groups, we further explore whether the backbone contains \emph{neuron-level} structures that are sensitive to knowledge conflicts.
We identify \emph{scenario-sensitive neurons} by applying per-neuron Welch's $t$-tests with Bonferroni correction to FFN intermediate activations under TN and FP inputs, and compare \method{} with the base model.
According to the results, we make several observations (Obs.) as follows.
\textbf{Obs.\,\ding{182}}
As shown in Figure~\ref{fig:neuron_sensitivity}(a), scenario-sensitive neurons already exist in the backbone but concentrate heavily in deep layers, with the late-layer average reaching 61.7\%, and only 0.8\% in early layers.
Qwen3-8B exhibits the same concentration, with 57.6\% in late layers and 0.1\% in early layers.
\textbf{Obs.\,\ding{183}}
Figure~\ref{fig:neuron_sensitivity}(b) further shows that \method{}'s learned participation policy aligns with this uneven landscape: the per-layer deviation of the participation profile from its default, $|\alpha^{(l)}{-}1|$, correlates strongly with the sensitive-neuron proportion, reaching $r{=}0.677$ on Qwen3-1.7B and $r{=}0.768$ on Qwen3-8B, both with $p{<}0.001$.
Layers rich in conflict-sensitive neurons exhibit larger deviations from the default participation profile, whereas layers with few such neurons remain nearly unchanged.
\textbf{Obs.\,\ding{184}}
Compared to the base model, \method{} slightly reduces the overall proportion of scenario-sensitive neurons by 1.7\% on Qwen3-1.7B and 0.5\% on Qwen3-8B, while the top-100 neurons by effect size exhibit higher Cohen's $|d|$ under \method{} on 1.7B, suggesting that the policy concentrates scenario-discriminative signals into fewer but more specialized neurons.
More details are provided in Appendix~\ref{app:neuron}.

\section{Conclusion}
In this paper, we introduced \method{}, a lightweight framework that mitigates knowledge conflicts in retrieval-augmented generation through adaptive parametric knowledge participation.
By learning a per-layer parametric knowledge participation vector via reinforcement learning, \method{} achieves more faithful knowledge arbitration between retrieved context and parametric memory while preserving general capabilities.
Comprehensive experiments across multiple benchmarks and LLMs show that \method{} outperforms competitive baselines and exhibits strong generalization, demonstrating its effectiveness and robustness across different settings.
These findings highlight the potential of \method{} to mitigate knowledge conflict in retrieval-augmented generation.

\noindent\textbf{Limitations.}
Although \method{} shows consistent gains on both Qwen and Llama backbones, our most comprehensive conflict-resolution comparisons are still concentrated on open-domain QA benchmarks such as MRQA and ConFiQA, and broader evaluation under noisier real-world retrieval pipelines remains future work.
In addition, the current design focuses on modulating FFN residual contributions, and jointly regulating attention pathways is a promising direction for further improving evidence aggregation under knowledge conflicts.
Finally, the learned participation policy depends on the constructed conflict supervision and reward design. Extending \method{} to more diverse conflict types, such as partial evidence or multi-hop inconsistency, remains an important direction for future research.

\section*{Ethics Statement}
Our research adheres to the ICLR Code of Ethics. All datasets used in this study are publicly
available. The code and related materials will be appropriately released to ensure transparency and reproducibility of our work.

\section*{Reproducibility Statement}
All information necessary to reproduce our results is documented in the Appendix, including dataset curation and preprocessing, training and inference hyperparameters, and so on.

\section*{AI Use Statement}
We used a large language model (LLM) solely for writing assistance—specifically, to correct grammar, improve wording, and enhance clarity. The LLM did not contribute to research design, data analysis, modeling, experiments, or interpretation of results. All technical content and conclusions were authored and verified by the authors, who take full responsibility for the manuscript.

\bibliographystyle{iclr2027_conference}
\bibliography{references}

\clearpage
\newpage
\appendix
\section{Appendix}

\subsection{License}
We are committed to ethical research practices and ensuring the reproducibility of our work. To this end, we present the licensing information for all datasets utilized in our experiments. These include Natural Questions (CC BY-SA 3.0), NewsQA (MIT License), SearchQA (Apache License 2.0), TriviaQA (Apache License 2.0), HotpotQA (CC BY-SA 4.0), and SQuAD (CC BY-SA 4.0). Our use of these datasets is in full compliance with their terms, as all aforementioned licenses expressly
permit their application in academic research.

\subsection{More Details about the Datasets in Evaluation}
\label{appendix:datasets}
In this section, we provide more details about the datasets evaluated in this paper as follows, with dataset statistics presented in Table~\ref{tab:data_statistics}. 


\subsubsection{In-Domain Datasets}
\begin{itemize}[leftmargin=0.5cm]
    \item \textbf{HotPotQA}~\citep{yang-etal-2018-hotpotqa} is a multi-hop question-answering dataset constructed from Wikipedia. 
    It requires models to reason over multiple supporting documents and identify evidence chains to answer complex questions.

    \item \textbf{NQ}~\citep{kwiatkowski-etal-2019-natural} is a large-scale open-domain question-answering dataset based on real queries issued to the Google search engine. 
    The answers are annotated as short spans or long passages from Wikipedia pages.

    \item \textbf{NewsQA}~\citep{trischler-etal-2017-newsqa} is a machine reading comprehension dataset built from CNN news articles. 
    Its questions are written by crowdworkers and require models to understand news passages and extract answers from the given articles.

    \item \textbf{SearchQA}~\citep{dunn2017searchqa} is an open-domain question-answering dataset constructed from Jeopardy! questions and search engine snippets. 
    It requires models to infer answers from noisy retrieved evidence rather than relying on a single clean reference passage.

    \item \textbf{SQuAD}~\citep{rajpurkar2016squad} is a reading comprehension dataset built from Wikipedia articles. 
    Each question is paired with a context paragraph, and the answer is typically annotated as a text span within the paragraph.

    \item \textbf{TriviaQA}~\citep{joshi-etal-2017-triviaqa} is a large-scale question-answering dataset containing trivia questions paired with independently collected evidence documents. 
    It is designed to evaluate whether models can answer challenging questions by reasoning over distant and potentially noisy evidence.
\end{itemize}

\subsubsection{Out-of-Domain Datasets}
\begin{itemize}[leftmargin=0.5cm]
    \item \textbf{ConfiQA}~\citep{bi2024contextdpoaligninglanguagemodels} is a benchmark designed to evaluate faithfulness in retrieval-augmented generation scenarios with knowledge conflicts. 
    It contains questions paired with provided contexts, where the model is expected to generate answers that are faithful to the given context rather than relying on its parametric knowledge.
\end{itemize}

\subsubsection{General Datasets}
\begin{itemize}[leftmargin=0.5cm]
    \item \textbf{MMLU}~\citep{hendrycks2021measuringmassivemultitasklanguage} is a multitask language understanding benchmark covering 57 subjects, including mathematics, history, computer science, law, and other academic or professional domains. 
    It is designed to evaluate models' broad world knowledge and problem-solving ability through multiple-choice questions.
\end{itemize}

\begin{table*}[h]
\centering
\caption{
\textbf{Detailed statistics of datasets used in evaluation.}
The first six test datasets are directly drawn from the MRQA Development Data. For ConFiQA, the reported number includes its Question Answering (QA), Multi-hop Reasoning (MR), and Multi-Conflicts (MC) subsets.
}
\small
\setlength{\tabcolsep}{6pt}
\renewcommand{\arraystretch}{1.15}
\resizebox{\textwidth}{!}{
\begin{tabular}{lccc}
\Xhline{1.2pt}
\rowcolor{tabletitle}
\textbf{Dataset} & \textbf{Category} & \textbf{Task Type} & \textbf{Test Samples} \\
\Xhline{1.2pt}

HotPotQA~\citep{yang-etal-2018-hotpotqa} 
& In-domain & Multi-hop QA & 5,901 \\

\rowcolor{gray!10}
NQ~\citep{kwiatkowski-etal-2019-natural} 
& In-domain & Open-domain QA & 12,836 \\

NewsQA~\citep{trischler-etal-2017-newsqa} 
& In-domain & Reading Comprehension & 4,212 \\

\rowcolor{gray!10}
SearchQA~\citep{dunn2017searchqa} 
& In-domain & Open-domain QA & 16,980 \\

SQuAD~\citep{rajpurkar2016squad} 
& In-domain & Reading Comprehension & 10,507 \\

\rowcolor{gray!10}
TriviaQA~\citep{joshi-etal-2017-triviaqa} 
& In-domain & Open-domain QA & 7,785 \\

ConFiQA~\citep{bi2024contextdpoaligninglanguagemodels} 
& Out-of-domain & Conflict QA & 18,000 \\

\rowcolor{gray!10}
MMLU~\citep{hendrycks2021measuringmassivemultitasklanguage} 
& General & Multi-task QA & 14,042 \\

\Xhline{1.2pt}
\end{tabular}
}
\label{tab:data_statistics}
\end{table*}


\subsection{More Details about the Training Data Construction}
\label{appendix:data_construction}
In this section, we provide details of the datasets used for training and validation.
Since the MRQA benchmark does not provide a standard validation split,
we first construct a contrastive dataset from the standard MRQA training sets~\citep{fisch-etal-2019-mrqa}.
We then randomly split the constructed dataset into training and validation sets,
with the corresponding dataset statistics presented in Table~\ref{tab:training_data_statistics}.
To enable LLMs to learn the adaptive knowledge participation policy through \method{},
we employ self-consistency to estimate the model's parametric belief and generate
two complementary types of examples.
Specifically, the context-following type refers to cases where the model's parametric knowledge is incorrect, and the gold context is provided, requiring the model to rely on external evidence.
In contrast, the context-resisting type refers to cases where the model's parametric knowledge is correct, but a conflicting context is provided, requiring the model to preserve reliable parametric knowledge against misleading evidence.
The specific construction procedure is detailed as follows.


\paragraph{Self-Consistency Filter.}
For each question $q_i$, we first estimate the model's dominant parametric belief without providing any retrieved context.
Following the self-consistency strategy~\citep{wang2022self, min2023beyond}, we prompt the model 5 times using only the question and collect its generated answers.
If a majority answer appears with a frequency of at least 3, we treat it as the model's dominant parametric answer and denote it as $\hat{r}_i^{p}$.
Questions without a stable majority answer are filtered out to ensure that the constructed conflict examples are based on a reliable estimate of the model's parametric belief~\citep{huang2025parammutesuppressingknowledgecriticalffns}.

\paragraph{Retrieval of Candidate Contexts.}
Following previous work~\citep{li2025matching, chen2026learning,zhang2025router,zhu2025large}, we adopt the standard retrieval pipeline implemented in FlashRAG~\citep{FlashRAG} to retrieve external passages for each question.
The retrieved passages serve as candidate contexts for constructing both supportive and conflicting evidence.
We denote the original supporting context associated with the gold answer as $c_i^{+}$.
To construct a counterfactual context, we select a passage from the retrieved candidates that is semantically similar to the query but does not contain the dominant parametric answer $\hat{r}_i^{p}$.
The selected passage is denoted as $c_i^{-}$.
This construction preserves topical relevance while introducing contextual evidence that conflicts with the model's parametric belief.

\paragraph{Construction of Contrastive Training Data.}
Based on the relationship between the dominant parametric answer $\hat{r}_i^{p}$ and the gold answer $r_i^{c}$, we construct two types of conflict examples.

\begin{itemize}[leftmargin=*]
    \item \textbf{Parametric Type.}
    If the dominant parametric answer $\hat{r}_i^{p}$ matches the gold answer $r_i^{c}$, the model's parametric knowledge is considered reliable.
    We pair the question $q_i$ with the counterfactual context $c_i^{-}$, which introduces misleading contextual evidence.
    The target answer is set to the parametric answer $\hat{r}_i^{p}$.
    This type of example requires the model to preserve correct parametric knowledge and avoid being misled by conflicting retrieved context.

    \item \textbf{Contextual Type.}
    If the dominant parametric answer $\hat{r}_i^{p}$ does not match the gold answer $r_i^{c}$, the model's parametric belief is considered unreliable.
    We pair the question $q_i$ with the supporting context $c_i^{+}$.
    The target answer is set to the dataset gold answer $r_i^{c}$.
    This type of example requires the model to override its incorrect parametric belief and ground its response in the provided context.
\end{itemize}

\paragraph{Training format.}
Each constructed example is converted into the training format used by \method{}.
For a constructed context $c_i$ and target answer $a_i$, we build the input prompt as
\begin{equation}
    x_i = \mathcal{T}(q_i,c_i),
\end{equation}
where $\mathcal{T}(\cdot)$ denotes the same prompt template used in the main experiments.
For Parametric Type examples, we set $c_i=c_i^{-}$ and $a_i=\hat{r}_i^{p}$.
For Contextual Type examples, we set $c_i=c_i^{+}$ and $a_i=r_i^{c}$.
The final training set is the union of the two constructed subsets:
\begin{equation}
    \mathcal{D}
    =
    \mathcal{D}_{p}
    \cup
    \mathcal{D}_{c},
\end{equation}
where each instance is represented as $(x_i,a_i)$.
This training data encourages the gate modules to learn both directions of knowledge arbitration: relying on parametric knowledge when it is correct and relying on contextual evidence when the parametric belief is incorrect.

\begin{table}[t]
\centering
\caption{
\textbf{Detailed statistics of the training and validation sets for different models.}
Each split contains context-following and context-resisting samples.
}
\label{tab:training_data_statistics}

\vspace{-0.3em}

\renewcommand{\arraystretch}{1.2}
\setlength{\tabcolsep}{6pt}

\begin{tabular}{clccc}
\Xhline{1.2pt}

\rowcolor{tabletitle}
\textbf{Model}
& \textbf{Split}
& \textbf{Context-Following}
& \textbf{Context-Resisting}
& \textbf{Total}
\\

\Xhline{1.2pt}

& Train
& 5021 {\footnotesize \textcolor{gray}{(53.38\%)}}
& 4386 {\footnotesize \textcolor{gray}{(46.62\%)}}
& 9407
\\

\rowcolor{gray!10}
\multirow{-2}{*}{Qwen3-1.7B}
& Valid
& 1255 {\footnotesize \textcolor{gray}{(53.38\%)}}
& 1096 {\footnotesize \textcolor{gray}{(46.62\%)}}
& 2351
\\

\hline

& Train
& 7216 {\footnotesize \textcolor{gray}{(50.59\%)}}
& 7049 {\footnotesize \textcolor{gray}{(49.41\%)}}
& 14265
\\

\rowcolor{gray!10}
\multirow{-2}{*}{Qwen3-8B}
& Valid
& 1803 {\footnotesize \textcolor{gray}{(50.58\%)}}
& 1762 {\footnotesize \textcolor{gray}{(49.42\%)}}
& 3565
\\

\Xhline{1.2pt}

\end{tabular}

\vspace{-5mm}
\end{table}

\subsection{More Details about the Baselines}
\label{appendix:baselines}
In this section, we report the details about the baselines used for comparison in this paper.

\begin{itemize}[leftmargin=0.5cm]
    \item \textbf{Vanilla-RAG}~\citep{ram2023incontextretrievalaugmentedlanguagemodels} is a standard retrieval-augmented generation baseline that directly prepends retrieved documents to the input prompt and generates answers conditioned on the augmented context.

    \item \textbf{CtrlA}~\citep{liu2024ctrlaadaptiveretrievalaugmentedgeneration} is a prompting-based method that adaptively controls the model's reliance on retrieved information during generation. 
    It aims to improve answer reliability by guiding the model to determine how much external context should be used.

    \item \textbf{CK-PLUG}~\citep{bi2025parametersvscontextfinegrained} is a decoding-based method for handling conflicts between contextual and parametric knowledge. 
    It introduces fine-grained control during decoding to better integrate retrieved evidence with the model's internal knowledge.

    \item \textbf{IRCAN}~\citep{shi2024ircanmitigatingknowledgeconflicts} identifies context-aware neurons through attribution analysis and reweights them to enhance the model's sensitivity to contextual knowledge.

    \item \textbf{SFT}~\citep{wei2022finetunedlanguagemodelszeroshot} is a supervised fine-tuning baseline that updates model parameters using labeled training examples. 
    It trains the model to generate target answers directly from the given query-context pairs.

    \item \textbf{Knowledgeable-R1}~\citep{lin2026resistingcontextualinterferencerag} is a fine-tuning-based method designed to improve robustness against contextual interference in retrieval-augmented generation. 
    It trains the model to better resist misleading retrieved contexts while preserving useful knowledge for answering questions. We utilize LoRA-based parameter-efficient fine-tuning and adhere to its standard training recipe to implement this baseline.

    \item\textbf{CoRect}~\citep{ma2026corectcontextawarelogitcontrast} is a training-free inference-time method that contrasts contextualized and context-free logits to localize conflict-inducing FFN layers and rectifies their hidden states to mitigate parametric suppression of retrieved evidence.

\end{itemize}

\begin{algorithm}[t]
\caption{\method{} Framework}
\label{alg:spark}
\SetKwInOut{Input}{Input}
\SetKwInOut{Output}{Output}

\Input{
Frozen backbone LLM $\pi_\theta$;
training dataset $\mathcal{D}=\{(x_i,a_i)\}_{i=1}^{N}$;
maximum number of iterations $N_{\mathrm{iter}}$;
group size $G$;
learning rate $\eta$.
}

\textcolor{linkgreen}{\texttt{// Stage 1: Adaptive FFN Participation Parameterization}}\;

Initialize the layer-wise participation parameters
$\psi=\{(\mathbf{w}_\ell,b_\ell)\}_{\ell=1}^{L}$
with $\mathbf{w}_\ell=\mathbf{0}$, $b_\ell=\mathrm{logit}(1/\lambda_\alpha)$
so that $\alpha_\ell=1$ initially (Eq.~\ref{eq:participation_initialization})\;

Insert the participation modules into the frozen backbone to
obtain the augmented policy $\pi_{\theta,\psi}$ using
Eqs.~\ref{eq:factor} and~\ref{eq:participation_ffn}\;

Keep the backbone parameters $\theta$ frozen throughout training\;

\textcolor{linkgreen}{\texttt{// Stage 2: Faithfulness-Guided Participation Policy Optimization}}\;

\For{$k \gets 1$ \KwTo $N_{\mathrm{iter}}$}{
    Sample a mini-batch $\mathcal{B}\subset\mathcal{D}$\;

    \ForEach{$(x_i,a_i)\in\mathcal{B}$}{
        Generate a group of $G$ candidate responses
        $\mathcal{O}_i=\{o_i^j\}_{j=1}^G \sim \pi_{\theta,\psi}(\cdot|x_i,c_i)$
        (Eq.~\ref{eq:group_generation})\;

        Compute the composite reward
        $r_i^j \gets R(o_i^j, a_i)$
        for each response $o_i^j\in\mathcal{O}_i$\;

        Estimate group-relative advantages from
        $\{r_i^j\}_{j=1}^{G}$\;
    }

    Compute the KL regularization $\mathcal{L}_{\mathrm{ref}}$
    (Eq.~\ref{eq:kl_ref})\;

    Compute the L2 regularization $\mathcal{R}_{\mathrm{regular}}$
    (Eq.~\ref{eq:participation_reg})\;

    Compute the overall objective
    $\mathcal{L} = -\mathcal{J}_{\mathrm{GRPO}}
    + \beta_{\mathrm{ref}}\mathcal{L}_{\mathrm{ref}}
    + \lambda_{\mathrm{regular}}\mathcal{R}_{\mathrm{regular}}$ \;

    Update only the participation parameters:
    $\psi \gets
    \operatorname{Adam}
    \bigl(\psi,\nabla_{\psi}\mathcal{L},\eta\bigr)$\;
}

\Output{Optimized participation policy parameters $\psi^{*}$.}
\end{algorithm}


\subsection{More Details about \method{}}
\label{appendix:method}
In this section, we provide additional details about the reinforcement learning procedure of \method{}, including the reward design and response parsing rule.

\paragraph{Reward design.}
Following recent work on rule-based rewards for RL-based LLM training~\citep{guo2025deepseek, zhang2025router}, we employ a composite reward
$R = R_{\mathrm{format}} + R_{\mathrm{faithful}}$ with two components.

To encourage adherence to the required structured output, we first define a
\textbf{format reward}:
\begin{equation}
    R_{\mathrm{format}} =
    \begin{cases}
        1.0, & \text{valid format with answer}, \\
        0.5, & \text{valid format without answer}, \\
        0.0, & \text{otherwise}.
    \end{cases}
    \label{eq:format_reward}
\end{equation}
Inspired by~\citep{guo2025deepseek, qian2025toolrlrewardtoollearning}, we assign
an intermediate score of $0.5$ to well-structured but incomplete responses,
which provides partial credit and yields a denser learning signal than a purely
binary alternative.

We then define a \textbf{faithfulness reward} to encourage the final answer to be consistent with the context-supported reference answer. 
In particular, the reference answers used in training are the correct evidence associated with each scenario. When the context is reliable, the reference is the context answer, and when the context is misleading, the reference is the parametric answer.
Let $\hat{a}_i^j$ denote the answer extracted from the answer field of the $j$-th sampled response
for input $x_i$, and let $\mathcal{A}_i$ denote the set of acceptable reference
answers. The faithfulness reward is computed as
\begin{equation}
\begin{aligned}
    R_{\mathrm{faithful}}
    =
    \mathbf{1}\bigg[
    &\max_{a \in \mathcal{A}_i}
    \mathrm{EM}\big(
        \mathrm{norm}(\hat{a}_i^j), \\
    &\mathrm{norm}(a)
    \big)
    = 1
    \bigg],
\end{aligned}
\label{eq:faithful_reward}
\end{equation}
where $\mathrm{norm}(\cdot)$ denotes the normalization function and
$\mathrm{EM}(\cdot,\cdot)$ denotes exact match after normalization. For datasets
with a single reference answer, we set $\mathcal{A}_i=\{a_i\}$. If the response
does not contain a valid answer field, we set $R_{\mathrm{faithful}}=0$.

\paragraph{Response parsing and normalization.}
We parse each generated response according to the required structured output
template. When a valid answer field is detected, its content is extracted as
$\hat{a}_i^j$ and used for reward computation. Otherwise, the extracted answer
is treated as empty and the response receives no faithfulness reward.
During reward computation, we apply the same normalization function
$\mathrm{norm}(\cdot)$ to both the extracted answer and the reference answers.
Specifically, we lowercase the text, remove punctuation and articles, and
collapse consecutive whitespace. This normalization prevents the reward from
penalizing superficial surface-form differences while preserving the criterion
that the generated answer should match the context-supported target.

\subsection{Hyperparameters}
\label{appendix:hyperparameters}
In this section, we provide the hyperparameter settings used in the training stage of \method{} to better ensure reproducibility, which are detailed in Table~\ref{tab:grpo_hyperparameters}. These training settings are the same for all the models used in our paper.

\begin{table}[t]
\centering
\caption{Hyperparameter settings used for training in \method{}.}
\small
\setlength{\tabcolsep}{6pt}
\renewcommand{\arraystretch}{1.1}

\begin{tabular}{lc}
\Xhline{1.2pt}
\rowcolor{tabletitle}
\textbf{Hyperparameter} & \textbf{Value} \\
\Xhline{1.2pt}

Training iterations & 150 \\
\rowcolor{gray!10}
Episodes per iteration & 64 \\
Per-device batch size & 4 \\
\rowcolor{gray!10}
Optimizer & AdamW \\
Learning rate & $1\times10^{-4}$ \\

\hline

\rowcolor{gray!10}
Generations per sample $G$ & 4 \\
KL coefficient & 0.001 \\
\rowcolor{gray!10}
Sampling temperature & 0.6 \\
Top-$p$ & 0.95 \\
\rowcolor{gray!10}
Top-$k$ & 20 \\

\hline

Gate reg. coefficient & 0.01 \\
\rowcolor{gray!10}
Trainable parameters & Gates only \\

\Xhline{1.2pt}
\end{tabular}

\label{tab:grpo_hyperparameters}
\end{table}


\subsection{Pseudocode of Proposed \method{}}
\label{appendix:algorithm}
In this section, we present the step-by-step training algorithm of our \method{} in Algorithm~\ref{alg:spark}, which is self-explanatory.

\begin{table*}[t!]
\centering
\vspace{-0.1em}
\caption{
\textbf{Performance comparison across three datasets.}
Qwen models are evaluated in non-thinking mode, and Llama models are instruction-tuned variants.
The \textbf{boldfaced} scores represent the \textbf{best} results.
}
\scriptsize
\setlength{\tabcolsep}{3.2pt}
\renewcommand{\arraystretch}{1.15}
\newcommand{\paramcell}[2]{\makecell[c]{#1\\[-0.15em]\scriptsize{(#2)}}}
\newcommand{\noparam}{\textemdash}
\resizebox{\textwidth}{!}{
\begin{tabular}{llc ccc ccc ccc ccc}
\Xhline{1.2pt}
\rowcolor{tabletitle}
&
&
& \multicolumn{3}{c}{\textbf{NQ}}
& \multicolumn{3}{c}{\textbf{TriviaQA}}
& \multicolumn{3}{c}{\textbf{SQuAD}}
& \multicolumn{3}{c}{\textbf{Avg.}} \\
\rowcolor{tabletitle}
\multirow{-2}{*}{\textbf{Model}}
& \multirow{-2}{*}{\textbf{Method}}
& \multirow{-2}{*}{\makecell[c]{\textbf{Train.}\\\textbf{Params}}}
& EM & ACC & F1
& EM & ACC & F1
& EM & ACC & F1
& EM & ACC & F1 \\
\Xhline{1.2pt}

\multicolumn{15}{c}{\textit{Qwen Models}} \\
\hline

\rowcolor{gray!10}
\multirow{2}{*}{\cellcolor{white}Qwen-3-0.6B}
& Vanilla-RAG
& N/A
& 40.63 & 51.34 & 53.56
& 46.72 & 57.60 & 54.11
& 46.91 & 63.32 & 59.30
& 44.75 & 57.42 & 55.66 \\
& \method{}
& \paramcell{28.7K}{0.0048\%}
& \textbf{47.20} & \textbf{58.19} & \textbf{62.83}
& \textbf{60.26} & \textbf{71.91} & \textbf{68.40}
& \textbf{58.50} & \textbf{72.92} & \textbf{71.63}
& \textbf{55.32} & \textbf{67.67} & \textbf{67.62} \\

\hline
\multicolumn{15}{c}{\textit{Llama Models}} \\
\hline

\rowcolor{gray!10}
\multirow{2}{*}{\cellcolor{white}Llama-3.2-1B}
& Vanilla-RAG
& N/A
& 10.95 & 14.97 & 14.90
& 24.37 & 32.23 & 28.90
& 16.76 & 30.25 & 23.80
& 17.36 & 25.82 & 22.53 \\
& \method{}
& \paramcell{32.8K}{0.0027\%}
& \textbf{34.96} & \textbf{46.14} & \textbf{48.74}
& \textbf{50.84} & \textbf{65.93} & \textbf{59.08}
& \textbf{38.75} & \textbf{69.54} & \textbf{54.94}
& \textbf{41.52} & \textbf{60.54} & \textbf{54.25} \\

\hline
\rowcolor{gray!10}
\multirow{2}{*}{\cellcolor{white}Llama-3.2-3B}
& Vanilla-RAG
& N/A
& 53.43 & \textbf{66.94} & 69.41
& 60.42 & 77.15 & 69.25
& 70.67 & \textbf{86.86} & 83.22
& 61.50 & 76.98 & 73.96 \\
& \method{}
& \paramcell{86.0K}{0.0027\%}
& \textbf{56.80} & 65.07 & \textbf{71.95}
& \textbf{71.83} & \textbf{81.91} & \textbf{78.98}
& \textbf{76.04} & 85.75 & \textbf{86.18}
& \textbf{68.22} & \textbf{77.58} & \textbf{79.04} \\

\hline
\rowcolor{gray!10}
\multirow{2}{*}{\cellcolor{white}Llama-3.1-8B}
& Vanilla-RAG
& N/A
& 52.23 & 67.60 & 68.78
& 64.75 & 81.03 & 73.83
& 67.95 & 87.71 & 81.92
& 61.64 & 78.78 & 74.84 \\
& \method{}
& \paramcell{131.1K}{0.0016\%}
& \textbf{53.54} & \textbf{68.18} & \textbf{70.59}
& \textbf{68.25} & \textbf{85.48} & \textbf{78.11}
& \textbf{69.64} & \textbf{89.30} & \textbf{84.33}
& \textbf{63.81} & \textbf{80.99} & \textbf{77.68} \\

\Xhline{1.2pt}
\end{tabular}
}
\vspace{-0.5em}
\label{tab:general}
\end{table*}

\begin{table}[t]
\centering
\caption{
\textbf{Performance comparison on the MMLU benchmark under the 5-shot setting. }
Hum. and Soc. denote Humanities and Social Sciences, respectively.
}
\small
\setlength{\tabcolsep}{3.0pt}
\renewcommand{\arraystretch}{1.15}

\begin{tabular}{llccccc}
\Xhline{1.2pt}
\rowcolor{tabletitle}
&
& \multicolumn{5}{c}{\textbf{MMLU}} \\
\rowcolor{tabletitle}
\multirow{-2}{*}{\textbf{Size}}
& \multirow{-2}{*}{\textbf{Method}}
& \textbf{Hum.}
& \textbf{Soc.}
& \textbf{STEM}
& \textbf{Other}
& \textbf{Avg.} \\
\Xhline{1.2pt}

\multicolumn{7}{c}{\textit{Qwen Models}} \\
\hline

\multirow{2}{*}{0.6B}
& Origin
& 40.98 & 54.31 & 45.64 & 51.14 & 47.19 \\
& \cellcolor{gray!10}\method{}
& \cellcolor{gray!10}41.38
& \cellcolor{gray!10}54.76
& \cellcolor{gray!10}44.72
& \cellcolor{gray!10}50.72
& \cellcolor{gray!10}47.13 \\

\hline
\multirow{2}{*}{1.7B}
& Origin
& 52.77 & 68.54 & 59.40 & 64.18 & 60.24 \\
& \cellcolor{gray!10}\method{}
& \cellcolor{gray!10}51.80
& \cellcolor{gray!10}67.92
& \cellcolor{gray!10}57.72
& \cellcolor{gray!10}62.54
& \cellcolor{gray!10}59.04 \\

\hline
\multirow{2}{*}{8B}
& Origin
& 66.18 & 83.82 & 75.42 & 78.24 & 74.79 \\
& \cellcolor{gray!10}\method{}
& \cellcolor{gray!10}66.06
& \cellcolor{gray!10}83.62
& \cellcolor{gray!10}75.39
& \cellcolor{gray!10}78.08
& \cellcolor{gray!10}74.66 \\

\hline
\multicolumn{7}{c}{\textit{Llama Models}} \\
\hline

\multirow{2}{*}{1B}
& Origin
& 42.85 & 51.06 & 39.84 & 52.14 & 46.03 \\
& \cellcolor{gray!10}\method{}
& \cellcolor{gray!10}42.98
& \cellcolor{gray!10}51.58
& \cellcolor{gray!10}40.34
& \cellcolor{gray!10}52.08
& \cellcolor{gray!10}46.28 \\

\hline
\multirow{2}{*}{3B}
& Origin
& 57.45 & 68.87 & 51.25 & 66.59 & 60.58 \\
& \cellcolor{gray!10}\method{}
& \cellcolor{gray!10}57.39
& \cellcolor{gray!10}69.52
& \cellcolor{gray!10}50.78
& \cellcolor{gray!10}66.85
& \cellcolor{gray!10}60.65 \\

\hline
\multirow{2}{*}{8B}
& Origin
& 64.29 & 77.25 & 60.20 & 73.29 & 68.20 \\
& \cellcolor{gray!10}\method{}
& \cellcolor{gray!10}64.57
& \cellcolor{gray!10}77.19
& \cellcolor{gray!10}59.91
& \cellcolor{gray!10}73.25
& \cellcolor{gray!10}68.21 \\

\Xhline{1.2pt}
\end{tabular}

\label{tab:more_mmlu}
\end{table}
\subsection{Extending \method{} to More LLMs}
\label{appendix:generalization}
In this section, we provide more experimental results of our proposed \method{} on more LLMs. These LLMs include Qwen3-0.6B, Llama3.2-1B, Llama3.2-3B, and Llama3.1-8B. 
To comprehensively evaluate the performance, we adopt three metrics, including Exact Match (EM), token-level F1 score (F1), and Accuracy (Acc).
As shown in Table~\ref{tab:general}, \method{} observes a consistent improvement across different families and sizes of LLMs, demonstrating the versatility and scalability of \method{} across a wide spectrum of model architectures and sizes. Furthermore, we extend the MMLU evaluation to these LLMs to examine whether \method{} consistently maintains the general capabilities of different model families and sizes reported in Table~\ref{tab:more_mmlu}.

\begin{figure}[t]
    \centering
    \begin{subfigure}[t]{0.31\linewidth}
        \centering
        \includegraphics[width=\linewidth]{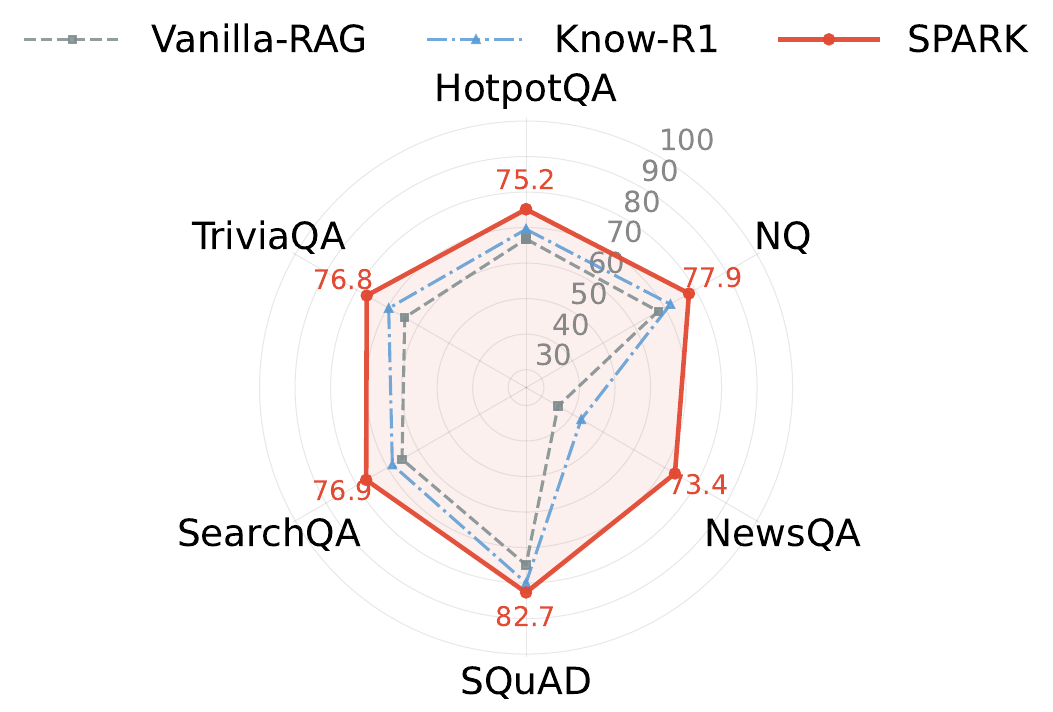}
        \caption{TP scenario: both parametric knowledge and context provide the correct answer.}
        \label{fig:method-a}
    \end{subfigure}
    \hfill
    \begin{subfigure}[t]{0.31\linewidth}
        \centering
        \includegraphics[width=\linewidth]{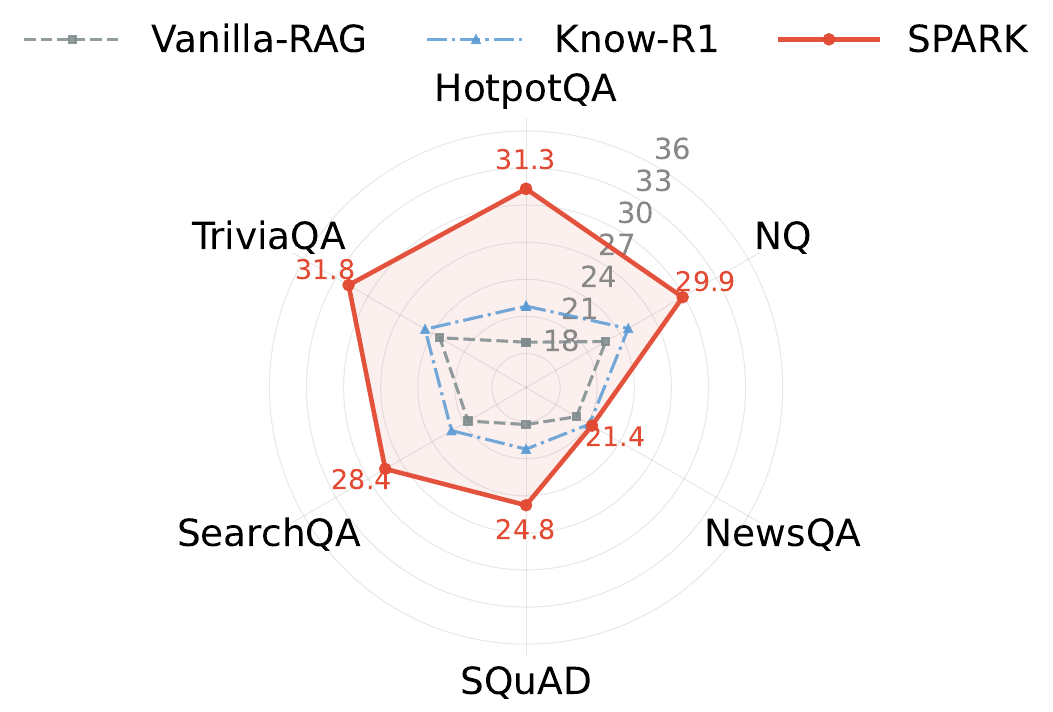}
        \caption{TN scenario: parametric knowledge is correct while context provides an incorrect answer.}
        \label{fig:method-b}
    \end{subfigure}
    \hfill
    \begin{subfigure}[t]{0.31\linewidth}
        \centering
        \includegraphics[width=\linewidth]{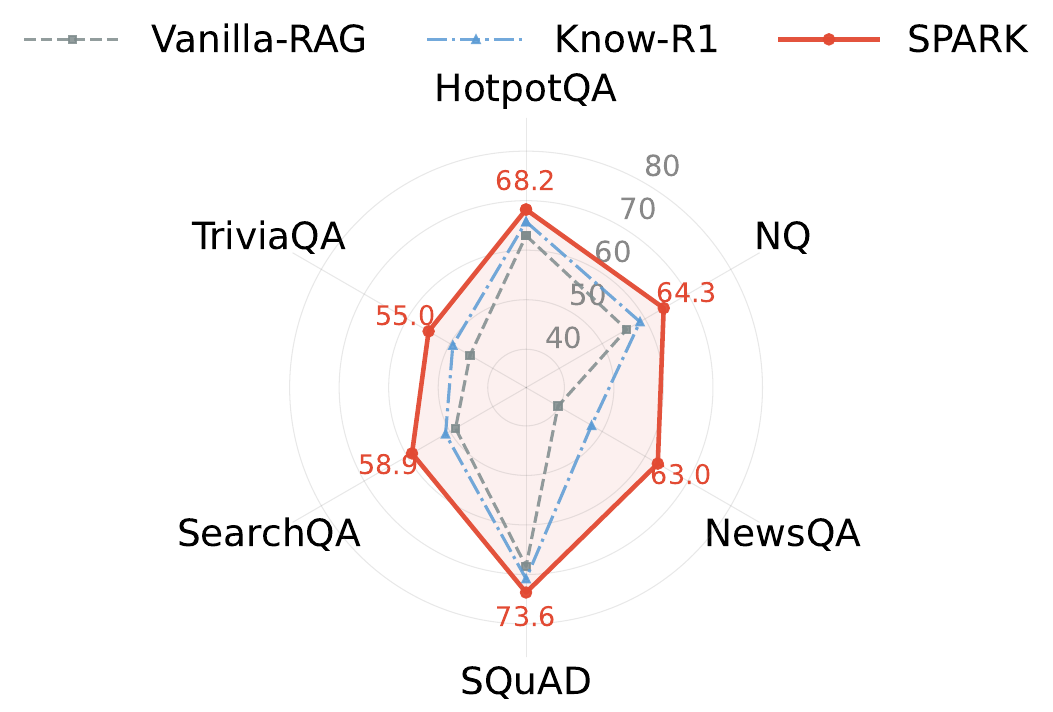}
        \caption{FP scenario: parametric knowledge is incorrect while context provides the correct answer.}
        \label{fig:method-c}
    \end{subfigure}
    \caption{Radar visualization of the participation vectors of Qwen3-1.7B under different knowledge conflict scenarios on the MRQA benchmark.}
    \label{fig:radar}
\end{figure}

\subsection{More Details about the scenario-specific Evaluation on MRQA Benchmark}
\label{appendix:scenarios}
In this section, we provide a more specific description of scenario-specific evaluation on the MRQA benchmark. First, we describe the three knowledge conflict scenarios used for evaluation and report the data statistics of each scenario. Then, we provide more experimental results using different LLMs.

\paragraph{Data Statistics.} To comprehensively evaluate the performance of \method{}, we define three typical knowledge conflict scenarios along two dimensions, including whether the model's parametric knowledge is correct and whether the context is right: \ding{182}~TP represents that both the parametric answer and the context agree on the correct answer. \ding{183}~TN represents that the parametric answer is correct, but the context is wrong. \ding{184}~FP represents that the parametric answer is wrong and the context provides the correct answer. 
Then, we utilize the same procedure described in Appendix~\ref{appendix:data_construction} to elicit the parametric answer of LLMs and to construct the conflict contexts. The corresponding data statistics for Qwen3-1.7B and Qwen3-8B are shown in Table~\ref{tab:scenario_data_statistics}.
We utilize the first two subsets for scenario-specific evaluation.

\begin{table}[t]
\centering
\caption{
\textbf{Detailed statistics of the MRQA scenario-specific split.}
Right-answer indicates that the LLMs could answer the question correctly.
Wrong-answer indicates that the LLMs could answer the question, but the answer is wrong.
No-answer indicates that the LLMs could not answer the question.
}
\label{tab:scenario_data_statistics}

\vspace{-0.3em}

\renewcommand{\arraystretch}{1.2}
\setlength{\tabcolsep}{4pt}

\begin{tabular}{clcccc}
\Xhline{1.2pt}

\rowcolor{tabletitle}
\textbf{Model}
& \textbf{Dataset}
& \textbf{Wrong-Answer}
& \textbf{Right-Answer}
& \textbf{No-Answer}
& \textbf{Total}
\\

\Xhline{1.2pt}


& HotpotQA
& 1088 {\footnotesize \textcolor{gray}{(72.15\%)}}
& 413 {\footnotesize \textcolor{gray}{(27.39\%)}}
& 7 {\footnotesize \textcolor{gray}{(0.46\%)}}
& 1508
\\

\rowcolor{gray!10}
& NQ
& 2286 {\footnotesize \textcolor{gray}{(63.15\%)}}
& 1333 {\footnotesize \textcolor{gray}{(36.82\%)}}
& 1 {\footnotesize \textcolor{gray}{(0.03\%)}}
& 3620
\\

& NewsQA
& 383 {\footnotesize \textcolor{gray}{(79.13\%)}}
& 101 {\footnotesize \textcolor{gray}{(20.87\%)}}
& 0 {\footnotesize \textcolor{gray}{(0.00\%)}}
& 484
\\

\rowcolor{gray!10}
& SQuAD
& 1905 {\footnotesize \textcolor{gray}{(68.70\%)}}
& 865 {\footnotesize \textcolor{gray}{(31.19\%)}}
& 3 {\footnotesize \textcolor{gray}{(0.11\%)}}
& 2773
\\

& SearchQA
& 2507 {\footnotesize \textcolor{gray}{(36.33\%)}}
& 4377 {\footnotesize \textcolor{gray}{(63.43\%)}}
& 16 {\footnotesize \textcolor{gray}{(0.23\%)}}
& 6900
\\

\rowcolor{gray!10}
& TriviaQA
& 1790 {\footnotesize \textcolor{gray}{(44.84\%)}}
& 2199 {\footnotesize \textcolor{gray}{(55.09\%)}}
& 3 {\footnotesize \textcolor{gray}{(0.08\%)}}
& 3992
\\

\hline

\multirow{-7}{*}{Qwen-3-1.7B}
& \textbf{Total}
& \textbf{9959} {\footnotesize \textcolor{gray}{(51.66\%)}}
& \textbf{9288} {\footnotesize \textcolor{gray}{(48.18\%)}}
& \textbf{30} {\footnotesize \textcolor{gray}{(0.16\%)}}
& \textbf{19277}
\\

\hline


& HotpotQA
& 876 {\footnotesize \textcolor{gray}{(58.09\%)}}
& 535 {\footnotesize \textcolor{gray}{(35.48\%)}}
& 97 {\footnotesize \textcolor{gray}{(6.43\%)}}
& 1508
\\

\rowcolor{gray!10}
& NQ
& 1857 {\footnotesize \textcolor{gray}{(51.30\%)}}
& 1696 {\footnotesize \textcolor{gray}{(46.85\%)}}
& 67 {\footnotesize \textcolor{gray}{(1.85\%)}}
& 3620
\\

& NewsQA
& 379 {\footnotesize \textcolor{gray}{(78.31\%)}}
& 95 {\footnotesize \textcolor{gray}{(19.63\%)}}
& 10 {\footnotesize \textcolor{gray}{(2.07\%)}}
& 484
\\

\rowcolor{gray!10}
& SQuAD
& 1634 {\footnotesize \textcolor{gray}{(58.93\%)}}
& 1045 {\footnotesize \textcolor{gray}{(37.68\%)}}
& 94 {\footnotesize \textcolor{gray}{(3.39\%)}}
& 2773
\\

& SearchQA
& 1192 {\footnotesize \textcolor{gray}{(17.28\%)}}
& 5574 {\footnotesize \textcolor{gray}{(80.78\%)}}
& 134 {\footnotesize \textcolor{gray}{(1.94\%)}}
& 6900
\\

\rowcolor{gray!10}
& TriviaQA
& 1100 {\footnotesize \textcolor{gray}{(27.56\%)}}
& 2836 {\footnotesize \textcolor{gray}{(71.04\%)}}
& 56 {\footnotesize \textcolor{gray}{(1.40\%)}}
& 3992
\\

\hline

\multirow{-7}{*}{Qwen-3-8B}
& \textbf{Total}
& \textbf{7038} {\footnotesize \textcolor{gray}{(36.51\%)}}
& \textbf{11781} {\footnotesize \textcolor{gray}{(61.11\%)}}
& \textbf{458} {\footnotesize \textcolor{gray}{(2.38\%)}}
& \textbf{19277}
\\

\Xhline{1.2pt}

\end{tabular}

\end{table}

\paragraph{Evaluation Results on more LLMs.} For a more comprehensive evaluation of \method{}, we evaluate the performance of Qwen3-1.7B illustrated in Figure~\ref{fig:radar} and Qwen3-8B on three different scenarios shown in Table~\ref{tab:scen-f1-all}.

\begin{table}[t]
\centering
\caption{
\textbf{F1 score of Qwen3-8B under three knowledge-conflict scenarios.}
The \textbf{boldfaced} scores represent the \textbf{best} results.
}
\label{tab:scen-f1-all}

\vspace{-0.3em}

\renewcommand{\arraystretch}{1.15}
\setlength{\tabcolsep}{2.5pt}

\begin{tabular}{clccccccc}
\Xhline{1.2pt}

\rowcolor{tabletitle}
\textbf{Scenario}
& \textbf{Method}
& \textbf{HotpotQA}
& \textbf{NQ}
& \textbf{NewsQA}
& \textbf{SQuAD}
& \textbf{SearchQA}
& \textbf{TriviaQA}
\\

\Xhline{1.2pt}

& Origin
& 84.76
& 83.78
& 81.52
& 89.53
& 85.42
& 83.53
\\

\rowcolor{gray!10}
& Know-R1
& \textbf{85.41}
& 84.19
& \textbf{81.78}
& 90.11
& 85.93
& 83.78
\\

\multirow{-3}{*}{\shortstack[l]{TP}}
& SPARK
& 84.11
& \textbf{84.84}
& 81.53
& \textbf{90.57}
& \textbf{86.48}
& \textbf{85.17}
\\

\hline

& Origin
& 39.81
& 42.78
& 36.77
& 30.85
& 44.37
& 48.48
\\

\rowcolor{gray!10}
& Know-R1
& 40.28
& 43.65
& 37.53
& 32.69
& 44.83
& 49.09
\\

\multirow{-3}{*}{\shortstack[l]{TN}}
& SPARK
& \textbf{52.11}
& \textbf{51.94}
& \textbf{44.98}
& \textbf{43.49}
& \textbf{56.73}
& \textbf{62.79}
\\

\hline

& Origin
& \textbf{74.85}
& \textbf{65.18}
& 68.48
& 80.55
& 59.77
& \textbf{47.94}
\\

\rowcolor{gray!10}
& Know-R1
& 73.76
& 65.01
& \textbf{68.97}
& 80.34
& \textbf{61.45}
& 46.65
\\

\multirow{-3}{*}{\shortstack[l]{FP}}
& SPARK
& 74.38
& 64.56
& 68.07
& \textbf{80.81}
& 61.03
& 47.78
\\

\Xhline{1.2pt}

\end{tabular}

\vspace{-5mm}
\end{table}

\begin{figure}[t]
    \centering
    \begin{subfigure}[t]{0.49\linewidth}
        \centering
        \includegraphics[width=\linewidth]{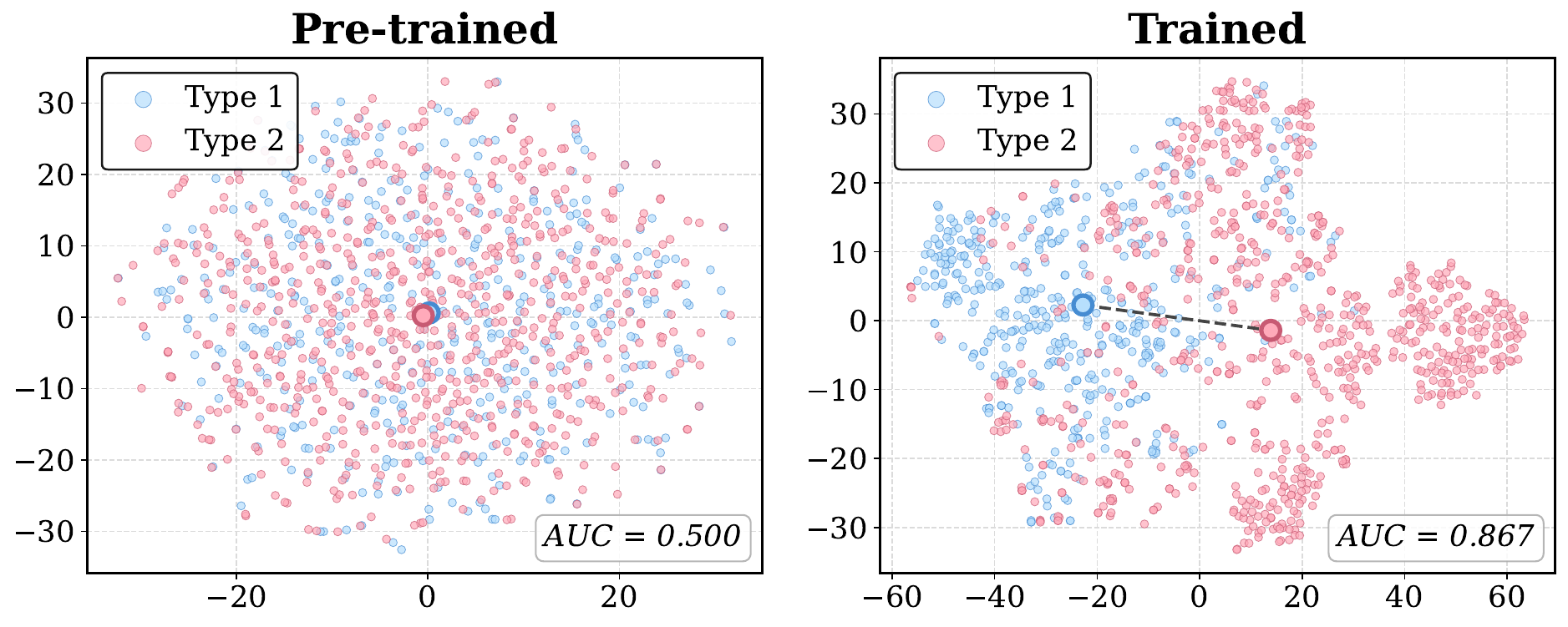}
        \caption{Visualization of the participation vector on Llama-3.2-1B.}
        \label{fig:method-a}
    \end{subfigure}
    \hfill
    \begin{subfigure}[t]{0.49\linewidth}
        \centering
        \includegraphics[width=\linewidth]{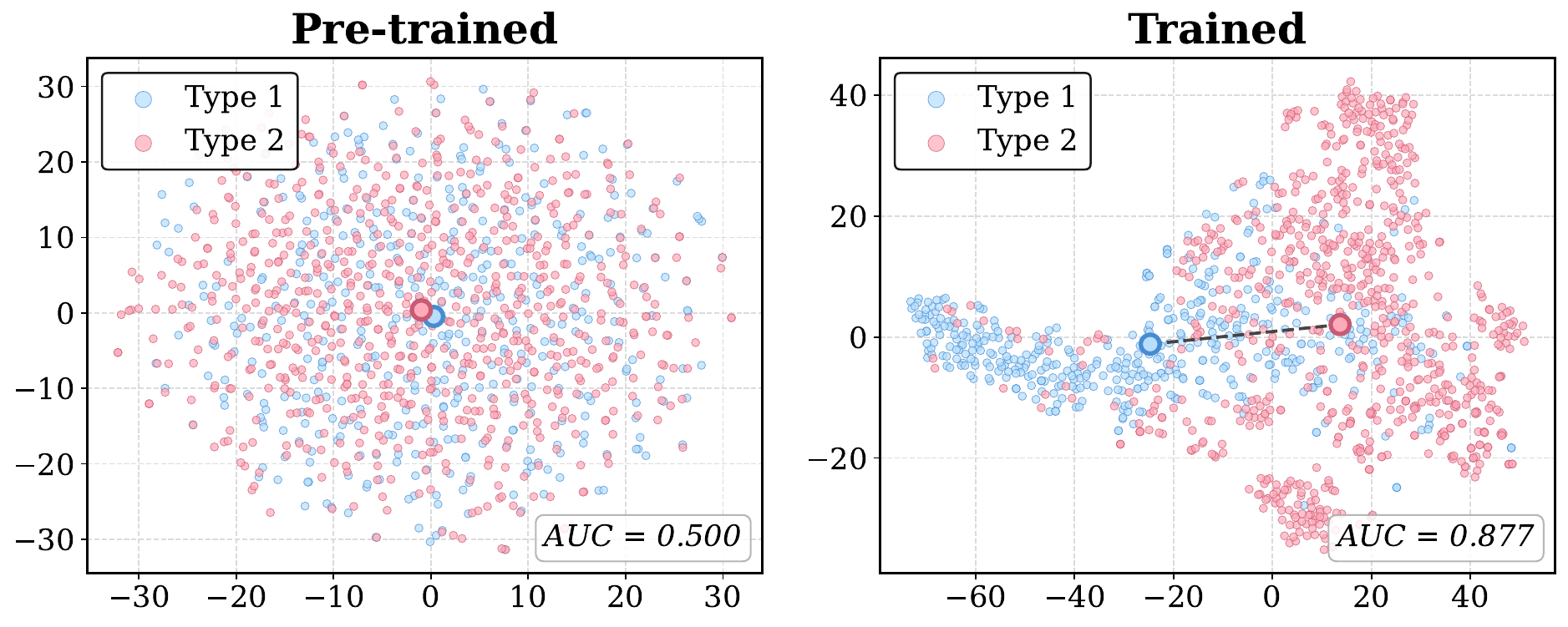}
        \caption{Visualization of the participation vector on Llama-3.2-3B.}
        \label{fig:method-b}
    \end{subfigure}
    \hfill
    \begin{subfigure}[t]{0.49\linewidth}
        \centering
        \includegraphics[width=\linewidth]{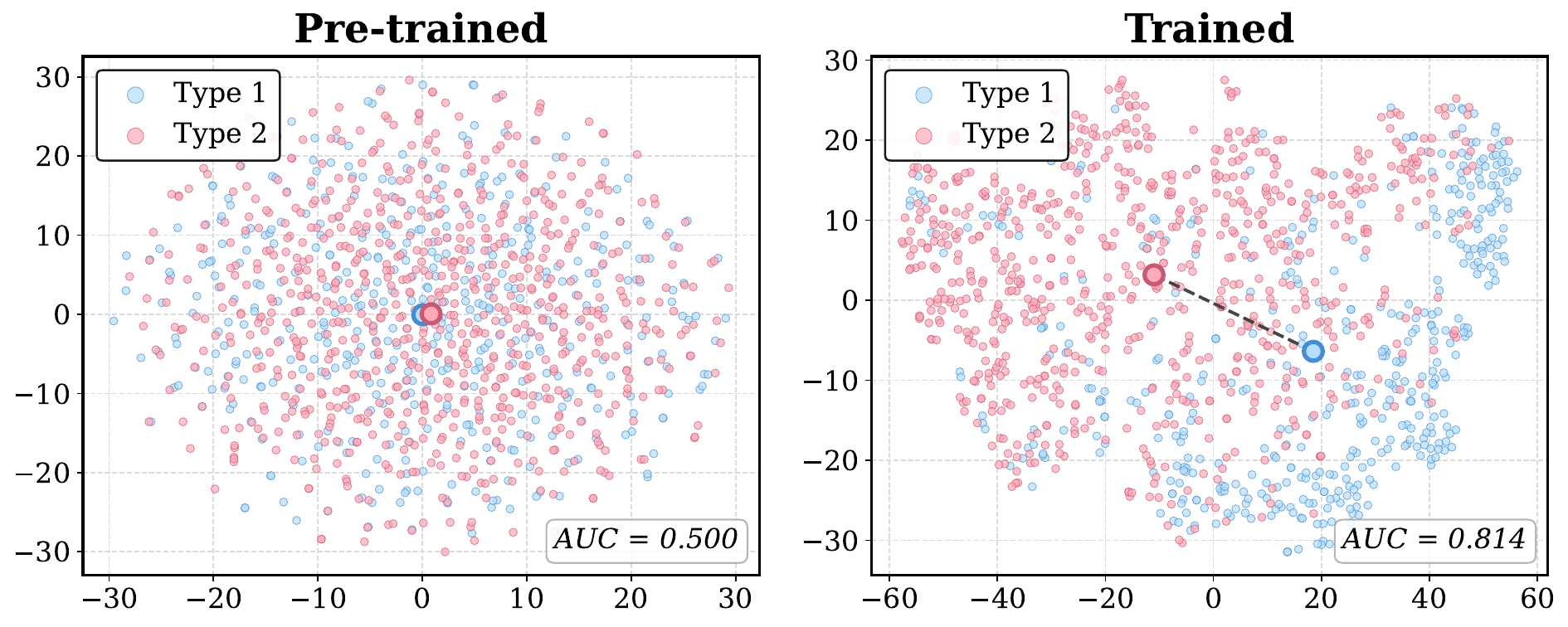}
        \caption{Visualization of the participation vector on Llama-3.1-8B.}
        \label{fig:method-b}
    \end{subfigure}
    \hfill
    \begin{subfigure}[t]{0.49\linewidth}
        \centering
        \includegraphics[width=\linewidth]{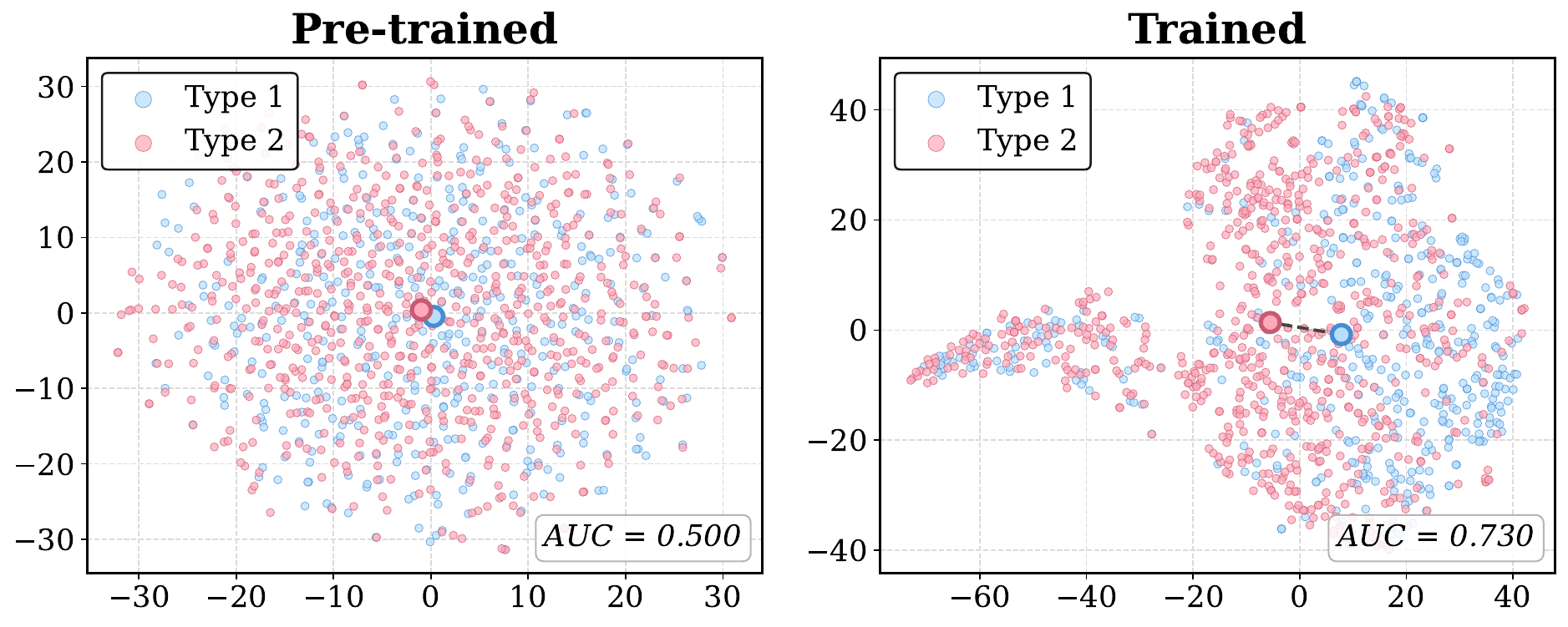}
        \caption{Visualization of the participation vector on Qwen-3-0.6B.}
        \label{fig:method-b}
    \end{subfigure}
    \hfill
    \begin{subfigure}[t]{0.49\linewidth}
        \centering
        \includegraphics[width=\linewidth]{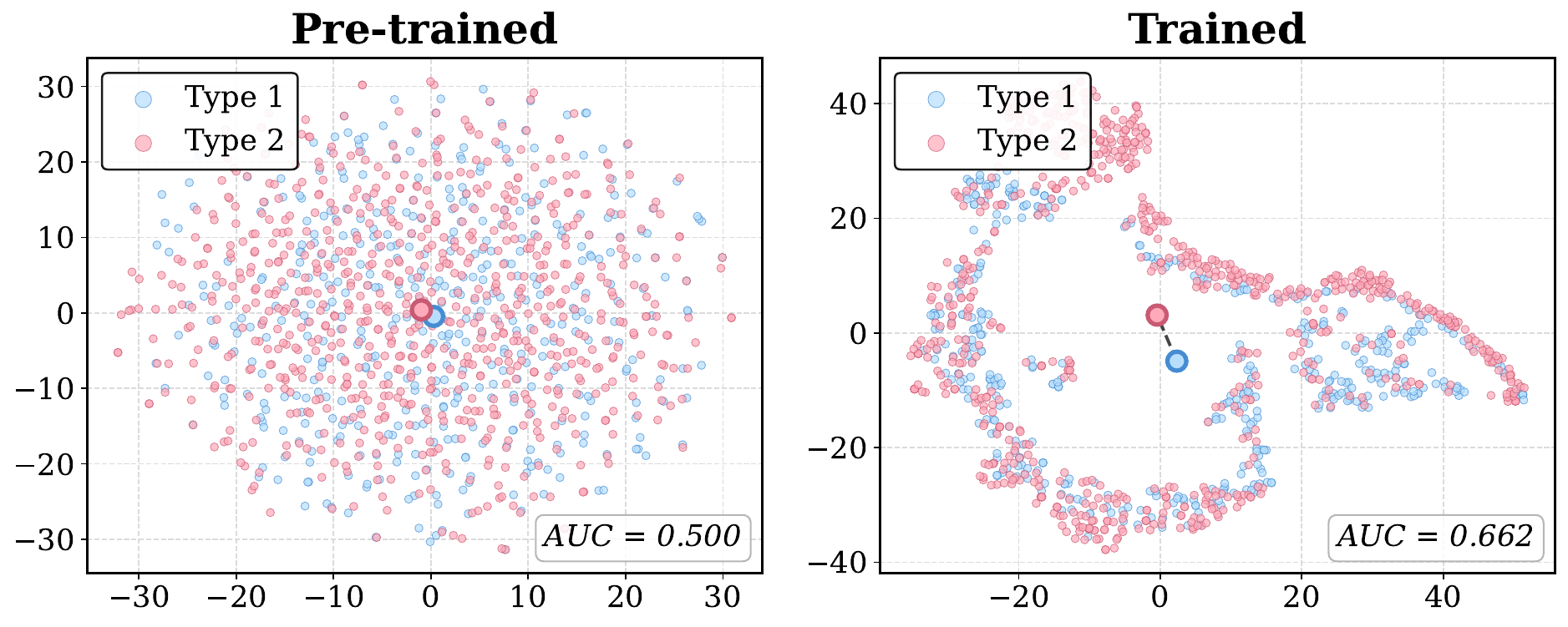}
        \caption{Visualization of the participation vector on Qwen-3-1.7B.}
        \label{fig:method-b}
    \end{subfigure}
    \hfill
    \begin{subfigure}[t]{0.49\linewidth}
        \centering
        \includegraphics[width=\linewidth]{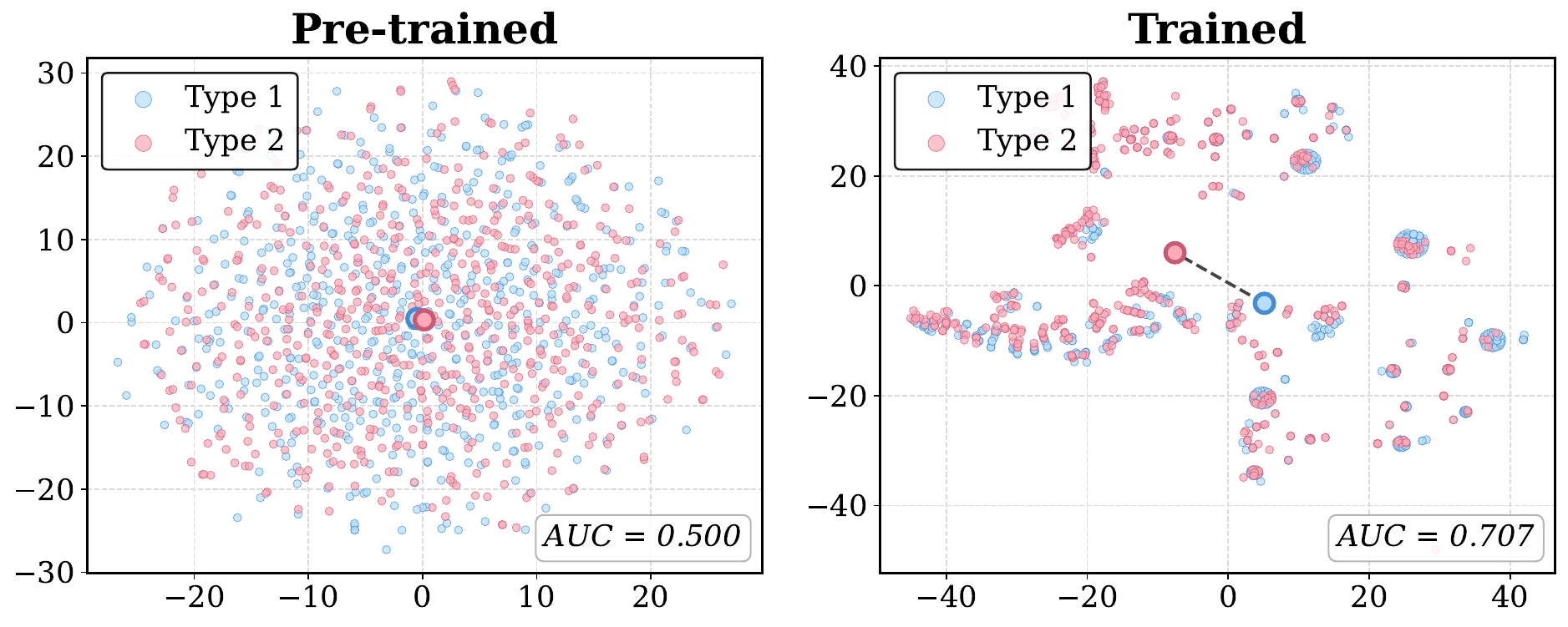}
        \caption{Visualization of the participation vector on Qwen-3-8B.}
        \label{fig:method-b}
    \end{subfigure}
    \vspace{-2mm}
    \caption{T-SNE visualization of the participation vectors of different LLMs.}
    \vspace{-1mm}
    \label{fig:t-sne}
\end{figure}

\subsection{T-SNE Visualization of Participation Vectors of \method{}}
\label{appendix:visiualization}
In this section, we exhibit the t-SNE visualizations of the participation vectors to further illustrate the effectiveness of our proposed \method{}. 
Specifically, following the data construction procedure detailed in Appendix~\ref{appendix:data_construction}, we sample 1,200 instances from MRQA evaluation subsets (200 per subset), categorized into two types: Type 1 (444 samples), where parametric knowledge is correct but the retrieved context is incorrect, and Type 2 (756 samples), where the retrieved context is correct but parametric knowledge is incorrect. 
For each instance, we collect the participation vectors from \method{} on Qwen-3-8B and represent them as an \(L\)-dimensional vector \(\mathbf{g}=[g_1, g_2, \ldots, g_L]\), where \(g_l\) denotes participation degree at layer \(l\).
As illustrated in Figure~\ref{fig:t-sne}, compared to the pre-trained model, the trained participation vectors exhibit visible separation between the two conflict types.
To quantify this, we further evaluate the linear separability of the gate vectors via 5-fold cross-validated logistic regression~\citep{kohavi1995study}. 
For example, the AUC of Qwen-3-8B increases from 0.500 to 0.707, suggesting that the gate module learns conflict-discriminative representations rather than applying uniform modulation across all inputs.
This further suggests that \method{} learns an effective participation policy, providing supporting evidence for its adaptive parametric knowledge participation behavior under different conflict scenarios.

\subsection{More Details about the Analysis of Scenario-Sensitive Neurons}
\label{app:neuron}

In this section, we provide the full experimental details and per-layer results on Qwen3-1.7B and Qwen3-8B for the neuron activation analysis discussed in
Section~\ref{sec:neuron_analysis}.
We first describe the experimental setup and the statistical procedure used to identify scenario-sensitive neurons. We then examine how these neurons concentrate in deeper layers and compare \method{} with the vanilla base model. Finally, we quantify the alignment between the learned participation policy and the layer-wise distribution of scenario-sensitive neurons.

\paragraph{Experimental Setup.}
We analyze the SwiGLU intermediate activations of each FFN layer at the last input token position, where the dimensionality is $d_m$. For each model, we collect activations on two conflict scenarios constructed
from six MRQA datasets:
\textbf{TN} represents that the parametric knowledge is correct, but the context is wrong, and \textbf{FP} represents that the parametric knowledge is wrong, but the context provides the gold answer.
For Qwen3-1.7B, which has $L{=}28$ layers and $d_m{=}6{,}144$, we obtain 9{,}288 TN and 9{,}959 FP samples. And for Qwen3-8B, which has $L{=}36$ layers and $d_m{=}12{,}288$, we obtain 11{,}781 TN and 7{,}038 FP samples. Both \method{} and the vanilla base model, which is equivalent to setting
$\alpha{=}1$ for all layers, are evaluated.

\paragraph{Identifying Scenario-sensitive Neurons.}
For each neuron $j$ at layer $l$, we perform a two-sided Welch's $t$-test~\citep{a967ba42-d0b9-3e65-976f-80cc6086b406} comparing its activation values across TN and FP samples.
To control the family-wise error rate over $d_m$ simultaneous tests per layer,
we apply Bonferroni correction with significance level
$\alpha_{\text{test}} = 0.05 / d_m$.
A neuron is deemed \emph{scenario-sensitive} if its corrected $p$-value falls
below this threshold.
We also compute Cohen's $d$ as a standardized effect size for each neuron.

\paragraph{Concentration in Deep Layers.}
Table~\ref{tab:neuron_concentration} reports the average proportion of
scenario-sensitive neurons across layer groups.
In both backbones, scenario-sensitive neurons are heavily concentrated in the
deeper layers, consistent with prior findings that mid-to-deep FFNs serve as
primary repositories of parametric
knowledge~\citep{dai2022knowledge}.

\begin{table}[h]
\centering
\caption{
\textbf{Average proportion of scenario-sensitive neurons (\%) by layer group
  in the \method{} model.}
}
\label{tab:neuron_concentration}

\vspace{-0.3em}

\renewcommand{\arraystretch}{1.2}
\setlength{\tabcolsep}{8pt}

\begin{tabular}{lccc}
\Xhline{1.2pt}

\rowcolor{tabletitle}
\textbf{Model}
& \textbf{Early}
& \textbf{Middle}
& \textbf{Late}
\\

\Xhline{1.2pt}

Qwen3-1.7B {\footnotesize \textcolor{gray}{(L0--9 / L10--17 / L18--27)}}
& 0.8
& 25.4
& 61.7
\\

\rowcolor{gray!10}
Qwen3-8B {\footnotesize \textcolor{gray}{(L0--11 / L12--23 / L24--35)}}
& 0.1
& 5.2
& 57.6
\\

\Xhline{1.2pt}

\end{tabular}

\vspace{-5mm}
\end{table}

\paragraph{Comparison between \method{} and the Base Model.}
Table~\ref{tab:neuron_spark_vs_base} presents the per-layer comparison between
\method{} and the base model.
Overall, \method{} slightly reduces the total proportion of scenario-sensitive
neurons, by 1.7 for 1.7B and 0.5 for 8B, while the top-100 neurons
ranked by effect size generally exhibit \emph{higher} Cohen's $d$ under
\method{} in the 1.7B model, suggesting that \method{}'s participation policy
concentrates scenario-discriminative signals into fewer but more specialized
neurons.

\begin{table}[h]
\centering
\caption{
\textbf{Scenario-sensitive neuron statistics: \method{} vs.\ Base.}
  ``\%sig'' denotes the proportion of neurons passing the Bonferroni-corrected
  threshold. ``$\Delta$pp'' denotes the difference in percentage points between \method{} and the base model. ``top$|d|$'' is the mean $|$Cohen's~$d|$ of the 100 neurons with the largest effect sizes. 
  Selected layers shown.
}
\label{tab:neuron_spark_vs_base}

\vspace{-0.3em}

\renewcommand{\arraystretch}{1.2}
\setlength{\tabcolsep}{4pt}

\begin{tabular}{r|rrrr|rrrr}
\Xhline{1.2pt}

\rowcolor{tabletitle}
 & \multicolumn{4}{c|}{\textbf{Qwen3-1.7B}} & \multicolumn{4}{c}{\textbf{Qwen3-8B}} \\

\rowcolor{tabletitle}
\textbf{Layer}
& \textbf{Base\%} & \textbf{SPARK\%} & \textbf{$\Delta$pp} & \textbf{top$|d|$}
& \textbf{Base\%} & \textbf{SPARK\%} & \textbf{$\Delta$pp} & \textbf{top$|d|$}
\\

\Xhline{1.2pt}

 0  &  0.0 &  0.0 & $+$0.0 & 1.53 &  0.0 &  0.0 & $+$0.0 & 1.17 \\
\rowcolor{gray!10}
 5  &  0.3 &  0.3 & $+$0.0 & 1.62 &  0.0 &  0.0 & $-$0.0 & 1.43 \\
10  &  4.4 &  4.0 & $-$0.4 & 1.61 &  0.2 &  0.2 & $+$0.0 & 1.53 \\
\rowcolor{gray!10}
12  & 12.0 & 14.0 & $+$2.0 & 1.62 &  0.3 &  0.3 & $+$0.0 & 1.50 \\
15  & 35.3 & 40.1 & $+$4.8 & 1.33 &  1.4 &  1.5 & $+$0.0 & 1.49 \\
\rowcolor{gray!10}
18  & 71.1 & 64.8 & $-$6.3 & 1.41 &  2.9 &  2.6 & $-$0.3 & 1.55 \\
20  & 73.5 & 66.3 & $-$7.2 & 1.31 &  7.6 &  7.1 & $-$0.5 & 1.61 \\
\rowcolor{gray!10}
21  & 70.4 & 60.9 & $-$9.5 & 1.26 & 11.3 & 11.2 & $-$0.0 & 1.66 \\
25  & 67.9 & 59.8 & $-$8.1 & 1.11 & 28.5 & 28.1 & $-$0.4 & 1.38 \\
\rowcolor{gray!10}
27  & 48.2 & 45.1 & $-$3.2 & 1.04 & 48.8 & 47.6 & $-$1.2 & 1.26 \\
30  &   -- &   -- &     -- &   -- & 74.4 & 70.2 & $-$4.2 & 1.07 \\
\rowcolor{gray!10}
35  &   -- &   -- &     -- &   -- & 59.6 & 62.5 & $+$2.9 & 0.99 \\

\hline

Avg & 31.2 & 29.6 & $-$1.7 &      & 21.4 & 20.9 & $-$0.5 &      \\

\Xhline{1.2pt}

\end{tabular}

\vspace{-5mm}
\end{table}

\paragraph{Correlation between Participation Policy and Sensitive-neuron
Distribution.}
To quantify how well the learned participation policy aligns with the
knowledge-conflict landscape, we compute the Pearson correlation between the
per-layer participation deviation $|\alpha^{(l)}{-}1|$ and the proportion of
scenario-sensitive neurons at layer $l$.
As shown in Table~\ref{tab:neuron_corr}, both models exhibit a strong positive
correlation, with $r{=}0.677$ for 1.7B and $r{=}0.768$ for 8B, both
significant at $p{<}0.001$, confirming that the RL-trained policy applies
stronger modulation at layers where knowledge-conflict signals are most
prevalent.

\begin{table}[h]
\centering
\caption{
\textbf{Pearson correlation between $|\alpha-1|$ and the proportion of
  scenario-sensitive neurons (\%sig).}
}
\label{tab:neuron_corr}

\vspace{-0.3em}

\renewcommand{\arraystretch}{1.2}
\setlength{\tabcolsep}{10pt}

\begin{tabular}{lcc}
\Xhline{1.2pt}

\rowcolor{tabletitle}
\textbf{Model}
& \textbf{$r$}
& \textbf{$p$-value}
\\

\Xhline{1.2pt}

Qwen3-1.7B & 0.677 & $<0.001$ \\

\rowcolor{gray!10}
Qwen3-8B   & 0.768 & $<0.001$ \\

\Xhline{1.2pt}

\end{tabular}

\vspace{-5mm}
\end{table}

\begin{table*}[t]
\caption{\label{tab:case_study}Case studies from Qwen3-1.7B under two knowledge-conflict scenarios.
Correct answers are highlighted in \textcolor[rgb]{0.7,0.3,0.3}{\textbf{Red}};
incorrect answers are highlighted in \textcolor[rgb]{0.13,0.67,0.8}{\textbf{Blue}}.
In TN, parametric knowledge is correct but the context is misleading;
in FP, parametric knowledge is wrong but the context provides the gold answer.}
\small
\resizebox{\linewidth}{!}{\begin{tabular}{p{0.98\linewidth}}
\Xhline{1.2pt}
\makecell[l]{%
\textbf{TN Case 1} \quad \textit{(HotpotQA --- Parametric knowledge: Manchester United)}\\
\textbf{Question:} Joe Ford played for which football club based out of Old Trafford?\\
\textbf{Context (cf):} Peter Ford (footballer) Peter Leslie Ford (born 10 August 1933) is an English former\\
footballer who played in the Football League for Port Vale and Stoke City \ldots\\
\\
\textbf{Vanilla-RAG:} \textcolor[rgb]{0.13,0.67,0.8}{\textbf{Unknown}} \quad
\textbf{SFT:} \textcolor[rgb]{0.13,0.67,0.8}{\textbf{West Ham United}} \quad
\textbf{\method{}:} \textcolor[rgb]{0.7,0.3,0.3}{\textbf{Manchester United}}
}
\tabularnewline
\hline
\makecell[l]{%
\textbf{TN Case 2} \quad \textit{(NewsQA --- Parametric knowledge: India)}\\
\textbf{Question:} What other country is communicating with the Tamil Tigers?\\
\textbf{Context (cf):} The Tamil Tigers are designated as a terrorist organization by 32 countries,\\
including the United States, Canada, and the member nations of the European Union \ldots\\
\\
\textbf{Vanilla-RAG:} \textcolor[rgb]{0.13,0.67,0.8}{\textbf{(empty)}} \quad
\textbf{SFT:} \textcolor[rgb]{0.13,0.67,0.8}{\textbf{Malaysia}} \quad
\textbf{\method{}:} \textcolor[rgb]{0.7,0.3,0.3}{\textbf{India}}
}
\tabularnewline
\Xhline{0.6pt}
\makecell[l]{%
\textbf{FP Case 1} \quad \textit{(HotpotQA --- Parametric knowledge: Western Cape)}\\
\textbf{Question:} The 1974 South African Grand Prix was a Formula One motor race held at a racing circuit\\
located in which South African Province?\\
\textbf{Context:} The 1974 South African Grand Prix was a Formula One motor race held at \textcolor[rgb]{0.7,0.3,0.3}{\textbf{Kyalami}} on\\
30 March 1974 \ldots Kyalami is a racing circuit located in Midrand, \textcolor[rgb]{0.7,0.3,0.3}{\textbf{Gauteng province}} \ldots\\
\\
\textbf{Vanilla-RAG:} \textcolor[rgb]{0.13,0.67,0.8}{\textbf{(empty)}} \quad
\textbf{SFT:} \textcolor[rgb]{0.13,0.67,0.8}{\textbf{Midrand}} \quad
\textbf{\method{}:} \textcolor[rgb]{0.7,0.3,0.3}{\textbf{Gauteng province}}
}
\tabularnewline
\hline
\makecell[l]{%
\textbf{FP Case 2} \quad \textit{(NaturalQuestions --- Parametric knowledge: Harrison Ford)}\\
\textbf{Question:} Who was originally cast to play Indiana Jones?\\
\textbf{Context:} Originally, Spielberg suggested Harrison Ford; Lucas resisted the idea \ldots\\
and did not want Ford to become his ``Bobby De Niro'' \ldots \textcolor[rgb]{0.7,0.3,0.3}{\textbf{Tom Selleck}} was cast\\
as Indiana Jones; however, CBS would not release Selleck from Magnum, P.I.~\ldots\\
\\
\textbf{Vanilla-RAG:} \textcolor[rgb]{0.13,0.67,0.8}{\textbf{Harrison Ford}} \quad
\textbf{SFT:} \textcolor[rgb]{0.13,0.67,0.8}{\textbf{Selleck}} \quad
\textbf{\method{}:} \textcolor[rgb]{0.7,0.3,0.3}{\textbf{Tom Selleck}}
}
\tabularnewline
\Xhline{1.2pt}
\end{tabular}}
\end{table*}

\begin{table*}[t]
\caption{\label{tab:case_study_8b}Case studies from Qwen3-8B under two knowledge-conflict scenarios.
Correct answers are highlighted in \textcolor[rgb]{0.7,0.3,0.3}{\textbf{Red}};
incorrect answers are highlighted in \textcolor[rgb]{0.13,0.67,0.8}{\textbf{Blue}}.
In TN, parametric knowledge is correct but the context is misleading;
in FP, parametric knowledge is wrong but the context provides the gold answer.}
\small
\resizebox{\linewidth}{!}{\begin{tabular}{p{0.98\linewidth}}
\Xhline{1.2pt}
\makecell[l]{%
\textbf{TN Case 1} \quad \textit{(HotpotQA --- Parametric knowledge: House of Commons)}\\
\textbf{Question:} The constituency which is represented by the 2015 elected Minister for Universities\\
and Science is represented in what wing of the UK parliament?\\
\textbf{Context (cf):} David Willetts David Linsay Willetts, Baron Willetts, (born 9 March 1956) is an English\\
Conservative Party politician, life peer, and academic. From 1992 to 2015, he was the Member of \ldots\\
\\
\textbf{Vanilla-RAG:} \textcolor[rgb]{0.13,0.67,0.8}{\textbf{Conservative}} \quad
\textbf{SFT:} \textcolor[rgb]{0.13,0.67,0.8}{\textbf{Conservative Party}} \quad
\textbf{\method{}:} \textcolor[rgb]{0.7,0.3,0.3}{\textbf{House of Commons}}
}
\tabularnewline
\hline
\makecell[l]{%
\textbf{TN Case 2} \quad \textit{(NaturalQuestions --- Parametric knowledge: Vicente Fox)}\\
\textbf{Question:} Who was elected president in Mexico in 2000?\\
\textbf{Context (cf):} \ldots a majority in both chambers of Congress---a situation that prevented Fox from\\
implementing his reform pledges. Nonetheless, the transfer of power in 2000 was quick and peaceful \ldots\\
\\
\textbf{Vanilla-RAG:} \textcolor[rgb]{0.13,0.67,0.8}{\textbf{Felipe Calder\'{o}n Hinojosa}} \quad
\textbf{SFT:} \textcolor[rgb]{0.13,0.67,0.8}{\textbf{Victor Manuel Mej\'{i}a}} \quad
\textbf{\method{}:} \textcolor[rgb]{0.7,0.3,0.3}{\textbf{Vicente Fox}}
}
\tabularnewline
\Xhline{0.6pt}
\makecell[l]{%
\textbf{FP Case 1} \quad \textit{(HotpotQA --- Parametric knowledge: Cruz Azul)}\\
\textbf{Question:} Rafael de Souza Pereira is a defensive midfielder for what organization that is the top\\
level of the Mexican football league system?\\
\textbf{Context:} Rafael Carioca Rafael de Souza Pereira (January 16, 1989) \ldots is a Brazilian footballer\\
who plays as a defensive midfielder for \textcolor[rgb]{0.7,0.3,0.3}{\textbf{Liga MX}} club Tigres UANL \ldots\\
\\
\textbf{Vanilla-RAG:} \textcolor[rgb]{0.13,0.67,0.8}{\textbf{Liga BBVA Bancomer}} \quad
\textbf{SFT:} \textcolor[rgb]{0.13,0.67,0.8}{\textbf{Liga BBVA Bancomer}} \quad
\textbf{\method{}:} \textcolor[rgb]{0.7,0.3,0.3}{\textbf{Liga MX}}
}
\tabularnewline
\hline
\makecell[l]{%
\textbf{FP Case 2} \quad \textit{(NaturalQuestions --- Parametric knowledge: Delhi, India)}\\
\textbf{Question:} Most polluted city in the world according to the World Health Organization?\\
\textbf{Context:} \ldots Position: 1, Country: Iran, Town/City: \textcolor[rgb]{0.7,0.3,0.3}{\textbf{Zabol}}, PM2.5: 217 \ldots\\
\\
\textbf{Vanilla-RAG:} \textcolor[rgb]{0.13,0.67,0.8}{\textbf{Not enough information provided}} \quad
\textbf{SFT:} \textcolor[rgb]{0.13,0.67,0.8}{\textbf{Dhaka}} \quad
\textbf{\method{}:} \textcolor[rgb]{0.7,0.3,0.3}{\textbf{Zabol}}
}
\tabularnewline
\Xhline{1.2pt}
\end{tabular}}
\end{table*}

\subsection{Case Studies}
\label{appendix:case}
In this section, we provide the qualitative case studies to illustrate the effectiveness of our \method{}.
Table~\ref{tab:case_study} and Table~\ref{tab:case_study_8b} present representative case studies from Qwen3-1.7B and Qwen3-8B, respectively, under the TN and FP knowledge-conflict scenarios.

In TN Case~1 of Qwen3-1.7B, the question asks which football club Joe Ford
played for, and the model's parametric knowledge correctly points to Manchester
United. However, the counterfactual context describes a different player
(Peter Ford) who played for entirely different clubs. Vanilla-RAG fails to
extract any relevant answer and outputs ``Unknown'', while SFT is misled into
producing ``West Ham United''. \method{} correctly retains its parametric
knowledge and answers ``Manchester United''. A similar pattern appears in TN
Case~2, where the counterfactual context omits the answer ``India'' entirely;
Vanilla-RAG produces an empty response, SFT hallucinates ``Malaysia'', and
only \method{} answers correctly.
These TN cases demonstrate that \method{}'s adaptive participation policy
enables the model to resist misleading context and preserve reliable
parametric knowledge when appropriate.
In FP Case~1, the model's parametric knowledge incorrectly recalls ``Western
Cape'' as the province where the 1974 South African Grand Prix took place,
while the context explicitly states Kyalami is located in Gauteng province.
SFT extracts the nearby but incorrect surface span ``Midrand'' (a city, not a
province), and Vanilla-RAG fails to produce any answer. \method{} correctly
follows the context and answers ``Gauteng province''. FP Case~2 is
particularly illustrative: the question asks who was originally cast to play
Indiana Jones. The context mentions both Harrison Ford (who was suggested by
Spielberg) and Tom Selleck (who was actually cast). Vanilla-RAG latches onto
the more salient name ``Harrison Ford''---which coincides with the model's
parametric memory---while SFT produces the incomplete answer ``Selleck''.
\method{} accurately extracts ``Tom Selleck'' from the context.
These FP cases show that \method{} can suppress incorrect parametric knowledge
and faithfully follow the retrieved context when it provides the correct
answer.

The Qwen3-8B cases in Table~\ref{tab:case_study_8b} exhibit consistent
patterns. Notably, in FP Case~2, the context contains a structured WHO table
listing Zabol (Iran) as the most polluted city, contradicting the parametric
answer ``Delhi, India''. Vanilla-RAG claims there is ``not enough information'',
SFT hallucinates ``Dhaka'', and only \method{} correctly extracts ``Zabol''
from the tabular data. Across both model scales, these cases illustrate that
\method{} achieves more faithful knowledge arbitration in both directions:
retaining parametric knowledge when the context is unreliable (TN) and
following contextual evidence when parametric knowledge is wrong (FP).



\subsection{Prompt Templates}
\label{appendix:prompt}
In this section, we provide the prompt templates used in our experiments, covering both training and evaluation.
For MRQA and ConFiQA benchmarks, we use the same unified instruction, without dataset-specific prompts.

\begin{figure*}[h]
\centering
\begin{tcolorbox}[
    title={Training Prompt},
    width=\textwidth,
    fonttitle=\bfseries,
    listing only,
    listing options={
        basicstyle=\ttfamily\small,
        breaklines=true,
        columns=fullflexible,
        literate={
            {<}{{\textless}}1
            {>}{{\textgreater}}1
        }
    }
]

\textbf{System Prompt:}
You are a helpful assistant who answers questions based on the provided background information.
The background information may be incorrect, so you should judge whether to believe yourself or to believe the background information.
You should think about the reasoning process and then provide the answer based on the given context.

\textbf{Prompt Template:}
Background:
{context}

Task Instruction:
Answer the question with the given background information above.
Provide your final answer in <answer> </answer> tags, for example <answer>Petri Alanko</answer>.

Q: {question}

A:
\end{tcolorbox}
\caption{Training prompt used for Qwen models in Non-Thinking Mode during GRPO training in \method{}.}
\label{fig:training_prompt_thinking}
\end{figure*}

\begin{figure*}[h]
\centering
\begin{tcolorbox}[
    title={Eval Prompt},
    width=\textwidth,
    fonttitle=\bfseries,
    listing only,
    listing options={
        basicstyle=\ttfamily\small,
        breaklines=true,
        columns=fullflexible
    }
]
\textbf{System Prompt:}
You are a helpful assistant who answers questions based on the provided background information.
The background information may be incorrect, so you should judge whether to believe yourself or to believe the background information.
You should think about the reasoning process and then provide the answer based on the given context.

\textbf{Prompt Template:}
Background:
{context}

Task Instruction:
Answer the question with the given background information above.
Provide your final answer in <answer> </answer> tags, for example <answer>Petri Alanko</answer>.

Q: {question}

A:
\end{tcolorbox}
\caption{Eval prompt on Qwen models in Non-Thinking Mode used for evaluation.}
\label{fig:eval_prompt_thinking}
\end{figure*}

\begin{figure*}[h]
\centering
\begin{tcolorbox}[
    title={Training Prompt},
    width=\textwidth,
    fonttitle=\bfseries,
    listing only,
    listing options={
        basicstyle=\ttfamily\small,
        breaklines=true,
        columns=fullflexible
    }
]
\textbf{System Prompt:}
You are a helpful assistant who answers questions based on the provided background information.
The background information may be incorrect, so you should judge whether to believe yourself or to believe the background information.
You should think about the reasoning process and then provide the answer based on the given context.

\textbf{Prompt Template:}
Background:
{context}

Task Instruction:
Answer the question with the given background information above.
Provide your final answer in <answer> </answer> tags, for example <answer>Petri Alanko</answer>.

Q: {question}

A:
\end{tcolorbox}
\caption{Training prompt used for Llama models during GRPO training in \method{}.}
\label{fig:training_prompt_thinking}
\end{figure*}

\begin{figure*}[h]
\centering
\begin{tcolorbox}[
    title={Eval Prompt},
    width=\textwidth,
    fonttitle=\bfseries,
    listing only,
    listing options={
        basicstyle=\ttfamily\small,
        breaklines=true,
        columns=fullflexible
    }
]
\textbf{System Prompt:}
You are a helpful assistant who answers questions based on the provided background information.
The background information may be incorrect, so you should judge whether to believe yourself or to believe the background information.
You should think about the reasoning process and then provide the answer based on the given context.

\textbf{Prompt Template:}
Background:
{context}

Task Instruction:
Answer the question with the given background information above.
Provide your final answer in <answer> </answer> tags, for example <answer>Petri Alanko</answer>.

Q: {question}

A:
\end{tcolorbox}
\caption{Eval prompt on Llama models used for evaluation.}
\label{fig:eval_prompt_thinking}
\end{figure*}




\end{document}